\documentclass[10pt,twocolumn,letterpaper]{article}

\usepackage{iccv}
\usepackage{times}
\usepackage{epsfig}
\usepackage{graphicx}
\usepackage{amsmath}
\usepackage{amssymb}
\usepackage[accsupp]{axessibility}
\usepackage{graphics} % for pdf, bitmapped graphics files
\usepackage{mathptmx} % assumes new font selection scheme installed
\usepackage{amsfonts} % blackboard math symbols
\usepackage{helvet}

\usepackage{url}
\usepackage{rotating}
\usepackage{xspace}
\usepackage{booktabs}       % professional-quality tables
\usepackage{nccmath}
\usepackage{nicefrac}       % compact symbols for 1/2, etc.
\usepackage{microtype}      % microtypography
\usepackage{comment}		
\usepackage{adjustbox}		
\usepackage{tabu}
\usepackage{calc}			
\usepackage{wrapfig}
\usepackage{tabularx}
\usepackage{array}
\usepackage{makecell}

\usepackage{multirow}
\usepackage{float}
\usepackage[hang,flushmargin]{footmisc}
\usepackage[neverdecrease]{paralist}
\usepackage{algorithm}
\usepackage{algorithmicx}
\usepackage{algpseudocode}

\usepackage{calc}
\usepackage{color}
\usepackage[table]{xcolor}
\usepackage{colortbl}
\usepackage{subfigure}
\usepackage{pifont}
\usepackage{animate}
\usepackage{nicefrac}       % compact symbols for 1/2, etc.
\usepackage{xfrac}

\usepackage[mathcal]{eucal}

\usepackage{ntheorem}
\newtheorem{assumption}{Assumption}

\usepackage{chapterbib}
\usepackage{standalone}
\usepackage[pagebackref=true,breaklinks=true,letterpaper=true,colorlinks,bookmarks=false,hypertexnames=false]{hyperref}

\newcommand{\figref}[1]{Fig.~\ref{#1}}
\newcommand{\tabref}[1]{Table~\ref{#1}}
\newcommand{\eqnref}[1]{Eq.~(\ref{#1})}
\newcommand{\secref}[1]{Sec.~\ref{#1}}
\newcommand{\calF}{{\mathcal{F}}}

\newcommand{\calL}{{\mathcal{L}}}

\newcommand{\be}{\begin{eqnarray}}
\newcommand{\ee}{\end{eqnarray}}
\newcommand{\bee}{\begin{eqnarray*}}
\newcommand{\eee}{\end{eqnarray*}}

\newcommand{\matrixb}{\left[ \begin{array}}
\newcommand{\matrixe}{\end{array} \right]}

\definecolor{CuGray}{gray}{0.9}
\newcolumntype{g}{>{\columncolor{CuGray}}c}

\newcommand{\cmark}{\ding{51}}%

\makeatletter
\DeclareRobustCommand\onedot{\futurelet\@let@token\@onedot}
\def\@onedot{\ifx\@let@token.\else.\null\fi\xspace}

\newcommand{\dashrule}[1][black]{%
  \color{#1}\rule[\dimexpr.5ex-.2pt]{4pt}{.4pt}\xleaders\hbox{\rule{4pt}{0pt}\rule[\dimexpr.5ex-.2pt]{4pt}{.4pt}}\hfill\kern0pt%
}

\newcommand*{\Scale}[2][4]{\scalebox{#1}{$#2$}}%

\def\eg{\emph{e.g}\onedot} 
\def\ie{\emph{i.e}\onedot} 
 
\def\etc{\emph{etc}\onedot} 
 
\def\etal{\emph{et al}\onedot}

\iccvfinalcopy % *** Uncomment this line for the final submission

\def\iccvPaperID{3833} % *** Enter the ICCV Paper ID here

\ificcvfinal\fi

\begin{document}

%%%%%%%%% TITLE
% \title{Self-supervised Monocular Depth and Motion Field Estimation via \linebreak Attentive and Contrastive Learning}
\title{Attentive and Contrastive Learning for Joint Depth and Motion Field Estimation}

\author{Seokju Lee \hspace{5 mm}
        Francois Rameau \hspace{5 mm}
        Fei Pan \hspace{5 mm}
        In So Kweon\\
        
        Korea Advanced Institute of Science and Technology (KAIST)\\
    
        {\tt\small \{seokju91,rameau.fr,feipan664\}@gmail.com, iskweon77@kaist.ac.kr}

}

\maketitle
\ificcvfinal\thispagestyle{empty}\fi

%%%%%%%%% ABSTRACT
\begin{abstract}
   Estimating the motion of the camera together with the 3D structure of the scene from a monocular vision system is a complex task that often relies on the so-called scene rigidity assumption. When observing a dynamic environment, this assumption is violated which leads to an ambiguity between the ego-motion of the camera and the motion of the objects. To solve this problem, we present a self-supervised learning framework for 3D object motion field estimation from monocular videos. Our contributions are two-fold. First, we propose a two-stage projection pipeline to explicitly disentangle the camera ego-motion and the object motions with dynamics attention module, called DAM. Specifically, we design an integrated motion model that estimates the motion of the camera and object in the first and second warping stages, respectively, controlled by the attention module through a shared motion encoder. Second, we propose an object motion field estimation through contrastive sample consensus, called CSAC, taking advantage of weak semantic prior (bounding box from an object detector) and geometric constraints (each object respects the rigid body motion model). Experiments on KITTI, Cityscapes, and Waymo Open Dataset demonstrate the relevance of our approach and show that our method outperforms state-of-the-art algorithms for the tasks of self-supervised monocular depth estimation, object motion segmentation, monocular scene flow estimation, and visual odometry.
\end{abstract}

%%%%%%%%%%%%%%%%%%%%%%%%%%%%%%%%%%%%%%%%%%%%%%%%%%%%%%%%%%%%%%%%%%%%%%%%%%%%%%%%
% Introduction
\section{Introduction}

\begin{figure}[t] 
    \centering
    {
    \includegraphics[width=0.98\linewidth]{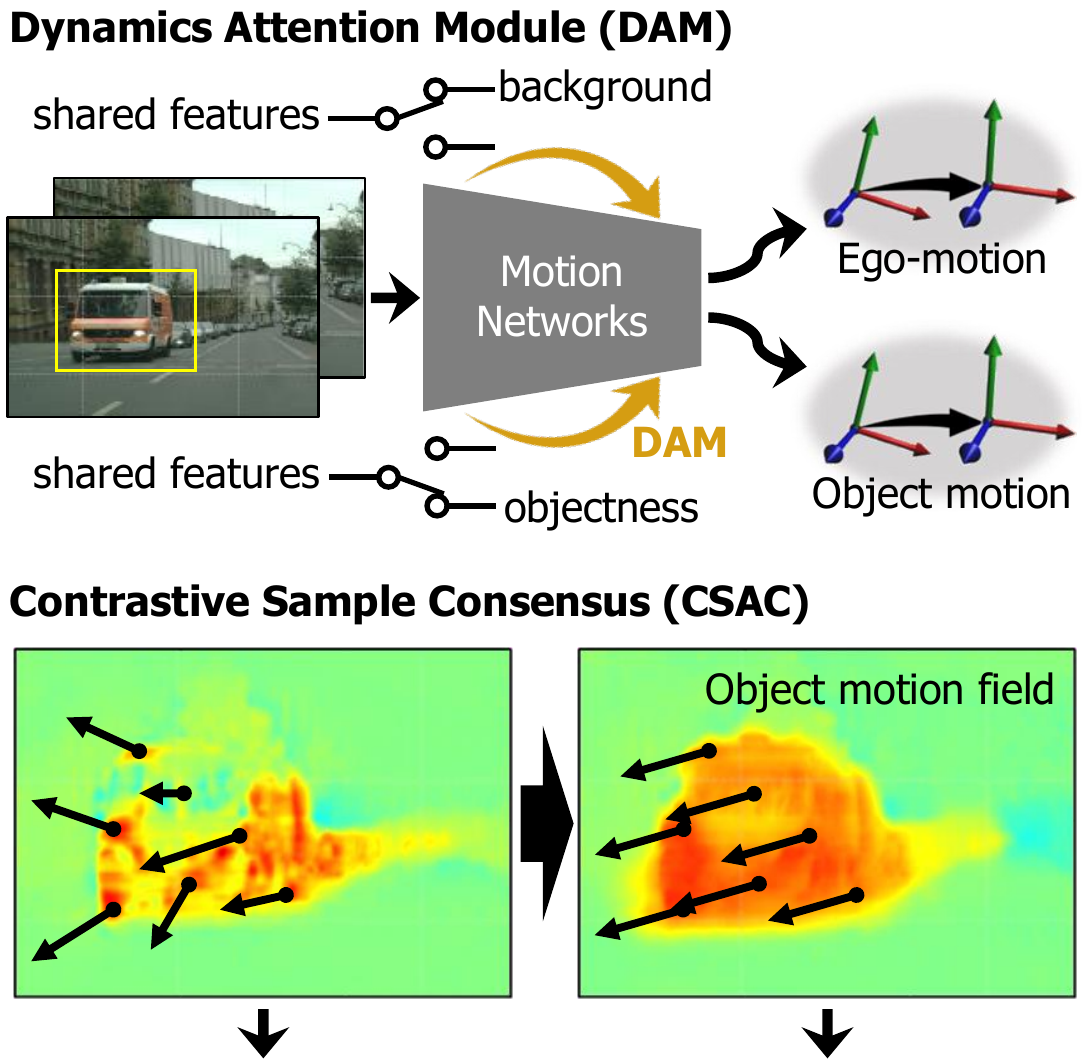}
    \subfigure[Baseline]{\label{teaser_a}
        \hspace{-0.4mm}
        \animategraphics[autoplay,loop,width=0.45\linewidth]{25}{./figure/aachen_066_0/}{0000}{0102}
        }
    \hspace{+1.2mm}
    \subfigure[Ours]{\label{teaser_b}
        \animategraphics[autoplay,loop,width=0.45\linewidth]{25}{./figure/aachen_066_1/}{0000}{0102}
        }
    }
    \vspace{-1mm}
    \caption{
    We introduce a unified motion modeling with our dynamics attention module (DAM) and a novel motion learning technique via contrastive sample consensus (CSAC). The last row shows synthesized views from the predicted depth and motion field. Compared to the baseline, our model learns the object motion fields in a more semantically plausible way, which enhances the distinction between the object and the background area. This is a \emph{video figure}, best viewed in \emph{Adobe Reader}.
    }
    \label{teaser}
    \vspace{-3mm}
\end{figure}

% 
%Simultaneous estimation of camera and scene geometry is the fundamental research topic in 3D computer vision and autonomous navigation for intelligent vehicle~\cite{Geiger2012CVPR}. 
The simultaneous estimation of the camera motion and scene geometry is a fundamental research topic in 3D computer vision.
% Simultaneous estimation of camera and scene geometry is the fundamental research topic in 3D computer vision.
% and autonomous navigation for intelligent vehicle~\cite{Geiger2012CVPR}. 
% Traditionally, this problem is tackled by feature-based methods~\cite{mur2015orb} that minimizes the geometric error across the feature correspondences, or the direct approaches~\cite{engel2014lsd} that minimizes the photometric inconsistency among warped adjacent frames.
Traditionally, this problem is tackled by feature-based methods~\cite{mur2015orb}, or direct approaches~\cite{engel2014lsd} that minimizes the photometric inconsistency among warped adjacent frames.
A pioneering work based on deep neural network (DNN) \cite{zhou2017unsupervised} uses the photometric error map as a self-supervisory signal to jointly train a depth and a motion network.
Inspired by this baseline structure, self-supervised depth, and motion learning framework has been widely studied~\cite{zhou2017unsupervised,li2017undeepvo,wang2018learning,mahjourian2018unsupervised,yang2018lego} with an additional self-supervisory signal such as geometric consistency~\cite{mahjourian2018unsupervised}, optical flow~\cite{ranjan2019competitive}, segmentation map~\cite{zhang2019dispsegnet}, edge and normal map~\cite{yang2018lego}.
%These photo-consistency-based optimization methods assume a static scene, where the 3D geometry is invariant in the world coordinates or the parts of moving objects are masked out to disregard the effect of non-rigid motion.
% These photo-consistency-based optimization methods assume a static scene, where the 3D geometry is invariant in the world coordinates or the parts of moving objects are masked out to disregard non-rigid motion.
These photo-consistency-based optimization methods assume a static scene or require to mask out moving objects to disregard non-rigid motions.
% Such works are designed to target the depth and ego-motion estimation, not dynamic motion estimation.
Such works aim at predicting the depth and ego-motion from a camera but are not suitable for dynamic scenes.

% Recently, the learning object motion along with depth and motion has gained interest to understand the dynamic scene environment~\cite{cao2019learning,lee2019learning,gordon2019depth,casser2019depth,casser2019unsupervised,klingner2020selfsupervised,dai2020self,brazil2020kinematic,lee2021learning}.
Recently, learning the objects' motion together with the camera's ego-motion and the depth has gain interest for dynamic scene understanding~\cite{cao2019learning,lee2019learning,gordon2019depth,casser2019depth,casser2019unsupervised,klingner2020selfsupervised,dai2020self,brazil2020kinematic,hur2020self,lee2021learning}.
We can distinguish mostly two types of approaches, namely the stereo-based and the monocular-based techniques.
The stereo-based techniques~\cite{cao2019learning,lee2019learning} take advantage of this sensor to disentangle the motion of static background and that of moving objects in the scene. 
%The stereo-based techniques~\cite{cao2019learning,lee2019learning} take advantage of the accurate depth prediction at metric scale that can be computed for each stereo frame. 
%This depth information coupled with semantic information (\ie, 2D object bounding box or instance segmentation) are used to disentangle the motion of static background and that of moving objects in the scene. 
%In the case of a monocular system, the ambiguity between the depth, ego-motion, and objects' motion becomes more complex to disentangle due to the impossibility to compute the metric depth for each frame. 
For monocular systems, the ambiguity between the depth, ego-motion, and objects' motion becomes more intricate due to the unavailability of metric depth for each frame. 
%Therefore, Monocular-based systems~\cite{casser2019depth,klingner2020selfsupervised,lee2021learning} mostly rely on instance segmentation labels to reduce the motion ambiguity between the camera and the objects' motion. 
Therefore, Monocular-based systems~\cite{casser2019depth,klingner2020selfsupervised,lee2021learning} rely on instance segmentation labels to reduce this ambiguity.
Despite compelling results, the need for highly expensive human-labeled data constitute an important limitation for their deployment and reduce the interest of the self-supervised depth and motion prediction framework.

% The stereo-based methods~\cite{cao2019learning,lee2019learning} predict depth maps from stereo images at each time step with rigidity assumption.
% Then, they use the scene geometry with 2D object bounding boxes~\cite{cao2019learning} to disentangle the motion of static background and that of moving objects in the scene. 
% In the case of a monocular system, the ambiguity of depth, ego-motion, and object motion becomes stronger than stereo setup due to the lack of rigidity.
% Monocular-based works \cite{casser2019depth,cao2019learning,klingner2020selfsupervised,lee2021learning} mostly leverage the instance segmentation labels to reduce the motion ambiguity between a camera and objects.
% Although they achieve competitive results to the stereo-based approaches but use highly expensive human-labeled data, which attenuates the strength of the self-supervised depth and motion prediction framework.

To reduce the data dependency problem and to offer more versatility, we propose a novel self-supervised learning framework for depth, camera motion, and object motion field estimation using weak semantic prior (\ie, 2D object bounding boxes) as illustrated in~\figref{teaser}.
% To tackle this problem, we design a self-supervised learning framework including depth, camera motion, and object motion field with weak semantic prior, 2D object box, in~\figref{teaser}.
% In contrast to the previous monocular-based approaches~\cite{casser2019depth,cao2019learning,klingner2020selfsupervised,lee2021learning}, we introduce a joint estimation framework of geometry in a dynamic scene with weaker prior knowledge. 
A major benefit of the proposed pipeline is that it helps to reduce the ambiguity between the camera's ego-motion and the objects' motion with cheaper data labels.
The distinctive points of our approach are summarized as follows:

\begin{list}{$\diamond$}{\leftmargin=4mm \itemindent=0mm \itemsep=0mm}
\item~We design a dynamics attention module that enables to train motion features dynamically when estimating the motion of a camera and objects through a two-stage projection.
We highlight that motion features can be efficiently extracted by disentangling dynamic objects and static backgrounds through the simple mechanism of attention modules within the shared motion encoder. 
\item~We propose a contrastive sample consensus for semantically plausible object motion field learning. Considering the rigid body characteristics of dynamic objects, we design a learning technique that effectively improves the capability to distinguish object's motion boundary. 
\item~We show that the proposed scheme achieves favorable results in motion segmentation, monocular depth and scene flow estimation, and visual odometry on the KITTI, Cityscapes, Waymo Open Dataset.
\end{list}

% We propose a hybrid\footnote{In this paper, the word ``hybrid'' has three meanings. First, we design the model to learn by using two heterogeneous prior knowledge, \emph{semantic} and \emph{geometric}. Second, our solution is motivated by the \emph{traditional} random sampling-based approach and \emph{recent} deep learning-based approach. Lastly, we design the proposed solver to operate using both \emph{CPU} and \emph{GPU} in consideration of computation time and memory efficiency.} solution for unsupervised 3D motion field regularization. With a given semantic prior knowledge of the 2D object box and its geometric prior knowledge of predicted depth, we design a differentiable regularization module combining traditional random sampling and recent deep learning approaches.

\begin{figure*}[t] 
	\centering
    \includegraphics[width=0.99\linewidth]{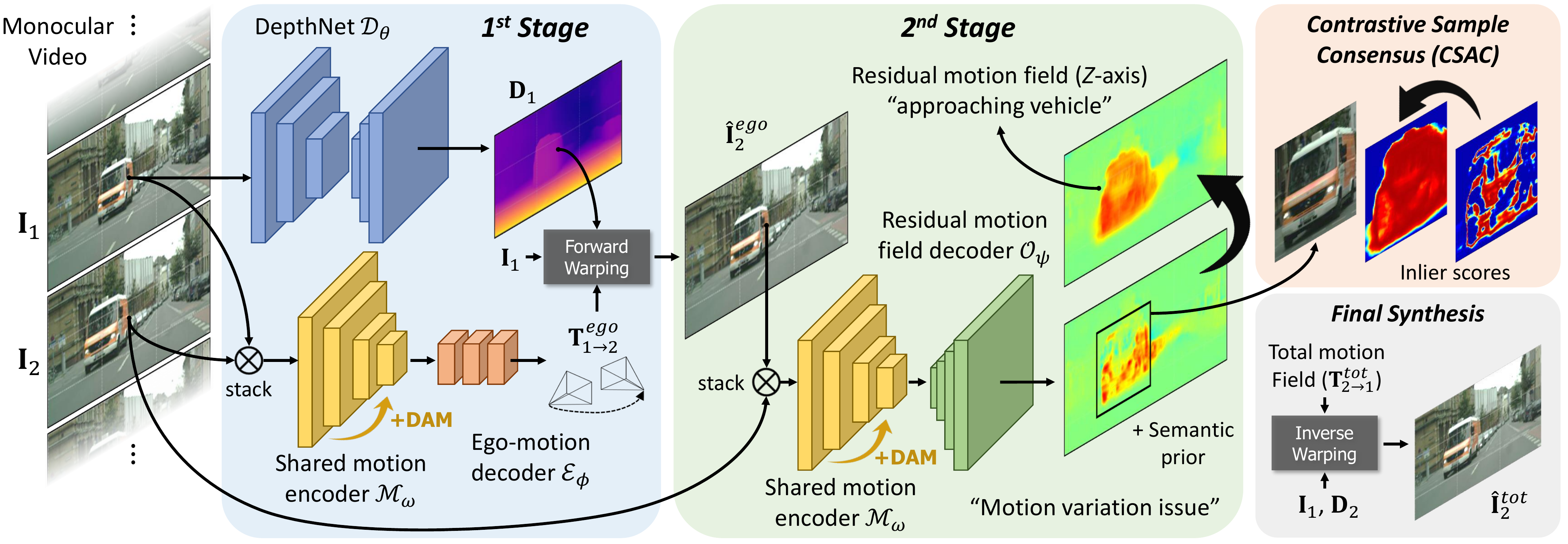}
	\caption{
	\textbf{Schematic overview of our self-supervised two-stage motion disentanglement.} First, we forward-warp $\mathbf{I}_1$ with the estimated camera motion and the depth map to synthesize $\hat{\mathbf{I}}_2^{ego}$. This ego-warped image is stacked with next frame, $\mathbf{I}_2$, and fed to the second stage projection to estimate the residual motion field. Finally, we generate a total composite motion field with the predicted ego-motion and residual motion field, and inverse-warp $\mathbf{I}_1$ to synthesize $\hat{\mathbf{I}}_2^{tot}$. The final synthesis is leveraged to optimize the networks as a self-supervisory signal.
	%Our motion encoder is shared while estimating the ego-motion and residual motion field. 
	Our motion encoder is shared to estimate the ego-motion and the residual motion field. 
	%Each motion requires features extracted on the background and object area, respectively, and this feature focusing is controlled by the proposed dynamics attention module (DAM).
	Each motion requires features extracted on the background and object area, respectively. This selective feature focusing is controlled by the proposed dynamics attention module (DAM).
	While learning the motion field, we propose contrastive sample consensus (CSAC) to solve the issue of local motion variation. Details of DAM and CSAC are elaborated in \secref{sec_3_2} and \secref{sec_3_3}.
	}
	\label{fig_overview}
\end{figure*}

% Related Works
\section{Related Works}

\subsection{Joint Training of Depth and Motion from Monocular Videos}

% Zhou~\etal~\cite{zhou2017unsupervised} first propose the joint depth and motion learning framework in a self-supervised manner by minimizing the photometric consistency across the monocular video.
Zhou~\etal~\cite{zhou2017unsupervised} first propose a self-supervised depth and motion framework minimizing the photometric consistency across a monocular video.
Following this publication, many improvements have been proposed ~\cite{wang2018learning,mahjourian2018unsupervised,godard2017unsupervised,yang2018lego,bian2019unsupervised,chen2019self,ranjan2019competitive,guizilini20203d,vasiljevic2020neural}.
% Following this publication, this field of research~\cite{wang2018learning,mahjourian2018unsupervised,godard2017unsupervised,yang2018lego,bian2019unsupervised,chen2019self,ranjan2019competitive,guizilini20203d,vasiljevic2020neural} has widely studied to infer both depth and ego-motion.
%Following the publication of this seminal work, this field of research~\cite{wang2018learning,mahjourian2018unsupervised,godard2017unsupervised,yang2018lego,bian2019unsupervised,chen2019self,ranjan2019competitive,guizilini20203d,vasiljevic2020neural} has widely studied to infer both depth and ego-motion.
% After the work, this field of research~\cite{wang2018learning,mahjourian2018unsupervised,godard2017unsupervised,yang2018lego,bian2019unsupervised,chen2019self,ranjan2019competitive,guizilini20203d,vasiljevic2020neural} has widely studied to infer both depth and ego-motion. 
%Wang \etal~\cite{wang2018learning} incorporate a second-order gradient descent-based pose refinement module, called direct visual odometry (DVO), into end-to-end training.
Wang \etal~\cite{wang2018learning} incorporate a second-order gradient descent-based pose refinement module, into end-to-end training.
Yang \etal~\cite{yang2018lego} introduce joint optimization of depth and motion with normal and edge information to force additional geometric loss while preserving the edges.
%Yang \etal~\cite{yang2018lego} introduce joint optimization of depth and motion with normal and edge information to force additional geometric loss while preserving the edges for training.
Mahjoiurian \etal~\cite{mahjourian2018unsupervised} enforce the geometric consistency across reconstructed 3D points as well as the photometric consistency.
Bian \etal~\cite{bian2019unsupervised} and Chen \etal~\cite{chen2019self} also impose the depth consistency loss by comparing multiple predicted depth, but they further estimate dynamic objects' mask~\cite{bian2019unsupervised} or camera intrinsics~\cite{chen2019self}.
Godard \etal~\cite{godard2019digging} propose a minimum reprojection loss to handle occlusion robustly and a multi-scale sampling method to reduce artifacts.
Ranjan \etal~\cite{ranjan2019competitive} introduce coordinated training frameworks composed of multiple neural networks for depth, camera motion, optical flow, and motion segmentation.
Guizilini~\etal~\cite{guizilini20203d} introduce a detail-preserving representation by learning representations that maximally propagate dense appearance and geometric information through 3D convolutions.
Vasiljevic~\etal~\cite{vasiljevic2020neural} represent the depth with differentiable pixel-wise projection rays for learning with uncalibrated single viewpoint cameras.
Recently, researches have been actively conducted to improve the performance of depth estimation in association with the semantic segmentation task. 
For instance, Klingner~\etal~\cite{klingner2020selfsupervised} leverage semantic segmentation guidance to adaptively mask out the photometric inconsistency in dynamic scenes.
Alternatively, Chen~\etal~\cite{chen2019towards} and Guizilini~\etal~\cite{guizilini2020semantically} improve the performance of monocular depth estimation while enhancing semantic understanding by extracting features that are commonly related to semantics and geometry.
They show that implicit feature learning through semantic prior knowledge can play an important role in 3D geometric perception.
% Zhang \etal~\cite{zhang2019dispsegnet} no motion

\subsection{Disentangling Camera and Object Motion}
% Learning Object Motion>
% cao2019learning
% gordon2019depth
% casser2019depth
% casser2019unsupervised
% klingner2020selfsupervised
% Self-supervised Object Motion and Depth Estimation from Video (CVPRw'20)
% Instance-wise Depth and Motion Learning from Monocular Videos (NIPSw'20)
% Kinematic 3D Object Detection in Monocular Video (ECCV'20)

Disentangling local object motion from the global camera ego-motion is a key to improve the robustness of both depth and motion estimation in dynamic situations.
Due to the motion ambiguity inherent to monocular-based techniques, most existing studies rely on stereo camera setup~\cite{cao2019learning,lee2019learning}.
This kind of system is advantageous in this context since metric scale depth (for each individual frame) coupled with semantic information (\eg, 2D objects' bounding boxes~\cite{cao2019learning} or segmentation labels~\cite{lee2019learning}) offers privileged information to disentangle the ambiguity between the static background and the moving objects.
While stereo-vision systems simplify the problem, solving this disentanglement using monocular cameras appears to be significantly more complex~\cite{gordon2019depth,casser2019depth,klingner2020selfsupervised,dai2020self,lee2021learning,brazil2020kinematic}.
% Due to the motion ambiguity inherent to monocular-based techniques, previous studies on the object motion estimation are tackled from a stereo camera setup~\cite{cao2019learning,lee2019learning} with rigidity assumption.
% They predict depth maps from stereo images at each time step, then use the information with 2D object bounding boxes~\cite{cao2019learning} or segmentation labels~\cite{lee2019learning} to disentangle the motion of static background and that of moving objects in the scene. 
% In the case of a monocular system, the ambiguity of depth, ego-motion, and object motion becomes stronger than stereo setup due to the lack of rigidity~\cite{gordon2019depth,casser2019depth,klingner2020selfsupervised,dai2020self,lee2021learning,brazil2020kinematic}.
%Some of the works~\cite{casser2019depth,lee2021learning} leverage instance segmentation map to identify the 2D pixel mask containing objects and estimation motion field of individual objects.
Some of the works~\cite{casser2019depth,lee2021learning} leverage instance segmentation map to estimate the motion field of individual objects.
%Then, they show the relative speed and moving direction of the objects in 3D space can be inferred from the motion field.
Casser~\etal~\cite{casser2019depth,casser2019unsupervised} especially focus on designing a geometric structure in the learning process by modeling the scene and objects. 
Lee~\etal~\cite{lee2021learning,lee2020instance} focus on a geometrically correct two-stage warping process that improves both photometric and geometric projection consistency in dynamic situations.
%Recently, studies have been introduced to disentangle the motion of objects with weaker semantic prior knowledge (\eg, from pixel-level to box-level prior). 
Recent works attempt to disentangle the motion of objects with weaker semantic prior knowledge (\eg, from pixel-level to box-level prior). 
Brazil~\etal~\cite{brazil2020kinematic} learn a 3D object bounding box with their orientation and 3D confidence from a monocular video.
Gordon~\etal~\cite{gordon2019depth} introduce a motion field representation with bounding box information to train depth, ego-motion, and dynamic motion from uncalibrated cameras.
Li~\etal~\cite{li2020unsupervised} extend the motion field representation with a motion sparsity loss without additional semantic prior knowledge.
Gao~\etal~\cite{gao2020attentional} propose attentional motion networks to adaptively focus on each object and background feature without semantic priors.

\section{Methodology}
\label{sec_3}
We introduce a two-stage pipeline for joint depth and motion learning. Our main objective is to disentangle the camera's and objects' motion in a self-supervised manner.
In this section, we present the two projection stages composing our system, and the networks: DepthNet, and MotionNet with a shared encoder and two branch decoders.
% In this section, we present the two projection stages composing our system (ego-motion and object-motion warping), as well as, the networks producing the outputs: DepthNet, and MotionNet with a shared encoder and two branch decoders.
% In this section, we present the two-stage projective geometry and the networks for each type of output: DepthNet, and MotionNet with a shared encoder and two branch decoders. 
%Further, we describe our novel dynamics attention module and contrastive sample consensus for efficient and semantics-guided representation of camera and object motions.
% Further, we detail our dynamics attention module and contrastive sample consensus for the semantics-guided representation of camera and object motions.
Further, we detail our dynamics attention module and contrastive sample consensus for modeling camera and object motions with the semantic guidance.

\subsection{Two-Stage Motion Disentanglement}
\label{sec_3_1}
The overall schematic framework of the proposed method is illustrated in~\figref{fig_overview}.
% In this process, we synthesize the nearby target frame $\hat{\mathbf{I}}_2$ to generate the self-supervisory signals by warping the source frame $\mathbf{I}_1$, where $\mathbf{I} \in \mathbb{R}^{3 \times H \times W}$ is an RGB image sampled from a monocular video.
The self-supervision of our architecture is achieved by warping the source frame $\mathbf{I}_1$ to its adjacent target frame $\hat{\mathbf{I}}_2$, where $\mathbf{I} \in \mathbb{R}^{3 \times H \times W}$ is an RGB image sampled from a monocular video.
The residual error resulting from this warping is used as a training signal.

\noindent\textbf{Stage 1 -- depth and ego-motion:}~We first predict each source and target view's depth map ($\mathbf{D}_1$, $\mathbf{D}_2$) via our DepthNet $\mathcal{D}_\theta: \mathbb{R}^{3 \times H \times W} \rightarrow \mathbb{R}^{1 \times H \times W}$ with trainable parameters $\theta$.
By concatenating two sequential images and depth maps ($\mathbf{I}_1$, $\mathbf{D}_1$, $\mathbf{I}_2$, $\mathbf{D}_2$) as an input, our proposed motion encoder $\mathcal{M}_\omega: \mathbb{R}^{8 \times H \times W} \rightarrow \mathbb{R}^{c_k \times h_k \times w_k}$ with trainable parameters $\omega$ extracts bottleneck motion features. Using the last layer's bottleneck feature as an input for the ego-motion decoder $\mathcal{E}_\phi: \mathbb{R}^{c_k \times h_k \times w_k} \rightarrow \mathbb{R}^6$ with trainable parameters $\phi$, we estimate the six-dimensional (three translations and Euler angles) relative transformation vector $\mathbf{T}^{ego}_{1 \rightarrow 2}$ as a forward camera ego-motion.
%We then synthesize the ego-warped image $\hat{\mathbf{I}}^{ego}_2$ and its depth map $ \hat{\mathbf{D}}^{ego}_2$ as outputs of the first projection stage as
We then synthesize the ego-warped image $\hat{\mathbf{I}}^{ego}_2$ and its depth map $ \hat{\mathbf{D}}^{ego}_2$ as outputs of the first stage as
\begin{equation}
\Scale[0.99]
{
\begin{aligned}
\{ \hat{\mathbf{I}}^{ego}_2, \hat{\mathbf{D}}^{ego}_2 \} = \mathcal{F}_{fwd} (\mathbf{I}_1, \mathbf{D}_1, \mathbf{T}^{ego}_{1 \rightarrow 2}, \mathbf{K}) ,
\end{aligned}
}
\label{eq_1}
\end{equation}
where $\mathcal{F}_{fwd}$ is a forward projection function proposed in \cite{lee2021learning}, and $\mathbf{K} \in \mathbb{R}^{3 \times 3}$ is a given camera intrinsic matrix.
We postulate that this ego-warped image and its projected depth are structurally aligned to the target view except for occluded and disoccluded regions if there are no moving objects. % in the scene. 

\noindent\textbf{Stage 2 -- residual motion field:}~In the second projection stage, we residually predict a motion translation field.
Since we synthesize the target view with the predicted camera motion, we conjecture that the misaligned regions are caused from local object motions. Using this clue, we model the object motion as a residual motion field $\mathbf{T}^{res}$ using our motion encoder and a residual motion field decoder $\mathcal{O}_\psi: \mathbb{R}^{c_k \times h_k \times w_k} \rightarrow \mathbb{R}^{3 \times H \times W}$ with trainable parameters $\psi$.
Specifically, we concatenate the outputs from the first projection stage and target frame's image and depth ($\hat{\mathbf{I}}^{ego}_2$, $\hat{\mathbf{D}}^{ego}_2$, $\mathbf{I}_2$, $\mathbf{D}_2$) as an input for the motion encoder. We again feed this to our motion encoder, and its output features are fed to the residual motion field decoder. This local object motion is represented only with a 3D translation field to reduce the rotation ambiguity from the camera.
Finally, we compose the total motion field $\mathbf{T}^{tot} \in \mathbb{R}^{6 \times H \times W}$ from the predicted ego-motion\footnote
{
Since the final warping is inverse direction, the direction of ego-motion also should be inverted, which is estimated by feeding a reverse-ordered input ($\mathbf{I}_2$, $\mathbf{D}_2$, $\mathbf{I}_1$, $\mathbf{D}_1$) to the ego-motion networks.
} 
and residual motion field through pixel-wise matrix multiplication.
Given this total motion field, source image, and target depth map, we synthesize the final target image and its depth map as
\begin{equation}
\Scale[0.99]
{
\begin{aligned}
\{ \hat{\mathbf{I}}^{tot}_2, \hat{\mathbf{D}}^{tot}_2 \} = \mathcal{F}_{inv} (\mathbf{I}_1, \mathbf{D}_2, \mathbf{T}^{tot}_{2 \rightarrow 1}, \mathbf{K}) ,
\end{aligned}
}
\label{eq_2}
\end{equation}
where $\mathcal{F}_{inv}$ is our pixel-wise inverse projection function.
We optimize the whole frameworks by minimizing the photometric and geometric errors between $\{ \mathbf{I}_2, \mathbf{D}_2 \}$ and $\{ \hat{\mathbf{I}}^{tot}_2, \hat{\mathbf{D}}^{tot}_2 \}$.
The loss functions will be discussed in \secref{sec_3_4}.

\begin{figure}[t] 
	\centering
    \includegraphics[width=0.99\linewidth]{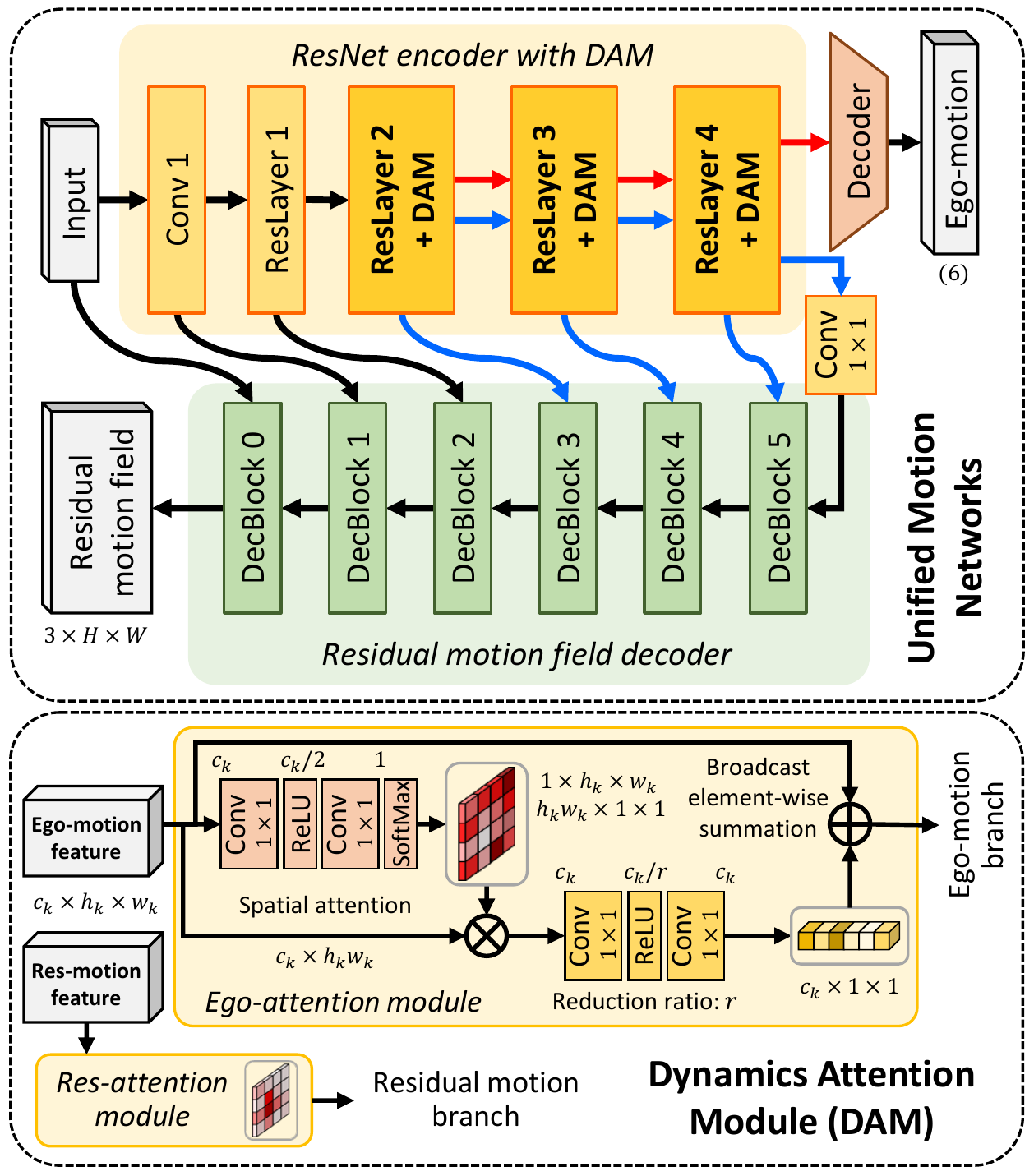}
	\caption{\textbf{Overall structure of unified motion networks with dynamics attention module (DAM).} Our motion networks consist of a shared motion encoder with an attention module for each residual layer (ResLayer), and two motion prediction branches: ego-motion decoder and residual motion (res-motion) field decoder. DAM has two self-attention modules to adaptively focus on background and local dynamic objects.}
	\label{dam}
\end{figure}

\begin{figure}[t]
	\begin{center}
		{\includegraphics[width=0.99\linewidth]{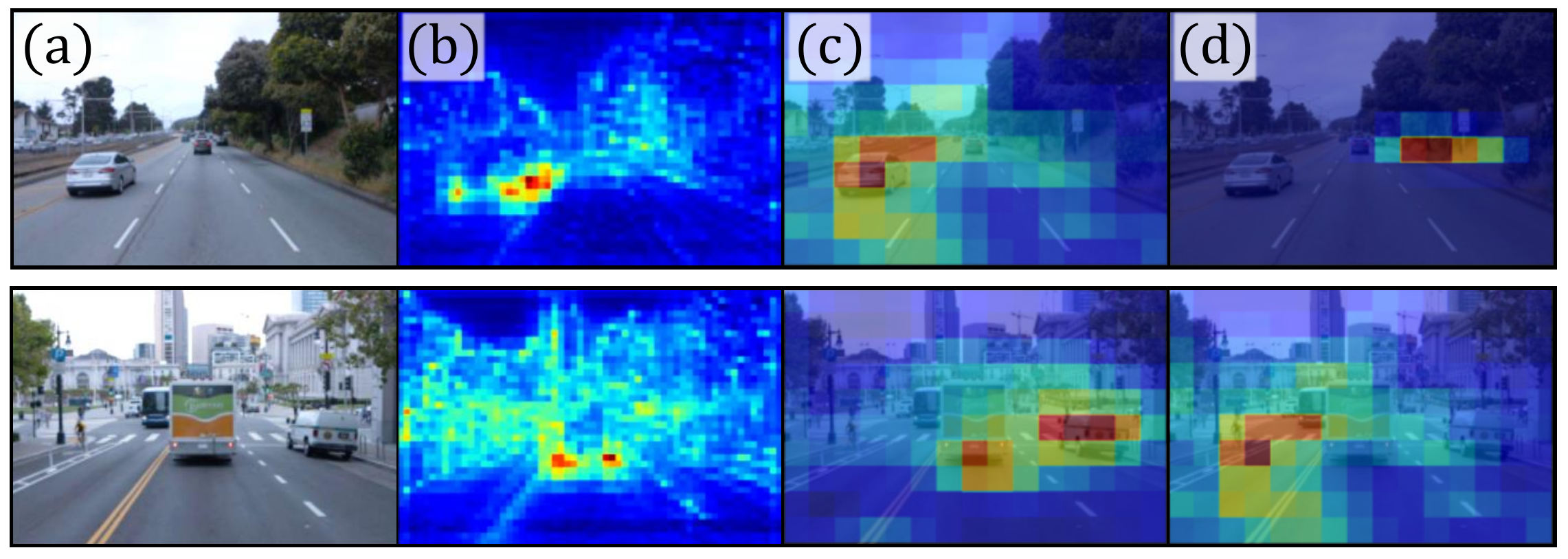}}\hspace{2mm}
	\end{center}
	\vspace{-0mm}
	\caption{
	\textbf{Qualitative results of attention maps.} (a) Input target images from Waymo Open Dataset. (b) Aggregated residual motion attentions on ResLayer-2 and -3 (mid-level). (c) and (d) residual motion and ego-motion attentions on ResLayer-4 (high-level, visually overlaid on the input image).
	}
	\label{rebut_attention}
	\vspace{-0mm}
\end{figure}

\subsection{Dynamics Attention Module}
\label{sec_3_2}
% While estimating camera motion in the first stage and residual motion field in the second stage, the motion encoder is designed to be shared.
% Although each operation is complementary to each other, we speculate that they have a common function, so we design it to be reused for efficiency and better motion feature representation. 
Since the camera motion and the residual motion field estimation are two complementary tasks, we postulate that using a shared encoder for these two tasks would improve their efficiency and the motion feature representation.
To maximize this effect, we propose \emph{dynamics attention module (DAM)} as described in \figref{dam}.
The encoding part of our unified motion networks is based on the ResNet-18~\cite{he2016deep} structure.
As an input for the networks, we concatenate two consecutive images and depth maps, which has eight channels in total. 
Motion features are learned while passing through each residual layer of the encoder. % can be omitted
In this process, we attach DAM after the residual layers (ResLayer-2, -3, and -4) to selectively extract the ego-motion and residual motion features.
We design DAM by referring to the generic self-attention structure that is transformed after context modeling introduced in GCNet~\cite{cao2019gcnet}. 
To be specific, we first squeeze the channel dimension with two $1 \times 1$ \emph{conv} layers and generate a spatial attention map via \emph{softmax} along the spatial dimension.
This spatial attention is multiplied to the input feature, which represents a global attention pooling for context embedding.
Then, with the reduction ratio set to $r = 4$, the pooled feature is transformed with a bottleneck of two $1 \times 1$ \emph{conv} layers.
Finally, the transformed feature is added to the input feature in element-wise for feature fusion.
This self-attention module is applied to each ego-motion and residual motion feature.
If the motion feature is extracted to predict the camera motion, the ego-motion attention module is activated, and if estimating the residual motion field, we operate the residual motion attention module.
Since the ego-motion and residual motion are in complementary relation, DAM enables selective motion focusing for each motion decoding as demonstrated in \figref{rebut_attention}.
% We demonstrate qualitative results of attention maps in \figref{rebut_attention}.
The ego-motion decoder is designed with four \emph{conv} layers to process the output feature of the last encoding layer.
The residual motion field decoder is composed of six decoding blocks (DecBlock).
Each decoding block aggregates output features of the bottom block and the encoding layer.
% More details of DAM and unified motion networks are described in the \texttt{supplement}.

\begin{figure}[t]
	\begin{center}
		{\includegraphics[width=0.99\linewidth]{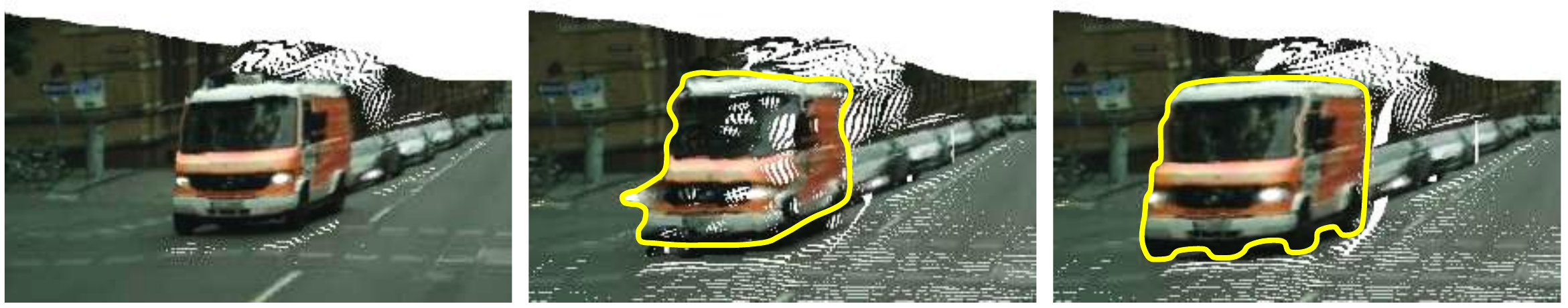}}\hspace{2mm}
	\end{center}
	\vspace{-0mm}
	\caption{
	\textbf{Left}: 3D point cloud visualization of $\mathbf{I}_1$ and $\mathbf{D}_1$. \textbf{Middle}: a motion variation issue occurs on view synthesis ($\mathbf{I}_1 \rightarrow \mathbf{I}_2$) with the baseline training. The networks try to minimize the photometric errors on headlight of the bus, while motions on homogeneous regions are not activated well. \textbf{Right}: Training the motion field with contrastive sample consensus (CSAC) to regularize the motion vectors for every pixel on the moving object. Yellow line indicates boundary of the object.
	}
	\label{fluctuation}
\end{figure}

\subsection{Contrastive Sample Consensus}
\label{sec_3_3}
\noindent\textbf{Motion variation issue}~
While learning the residual motion field, our motion networks are trained to minimize the local errors triggered by an individual object motion.
However, due to the limitation of self-supervisory optimization by photometric %consistency, an issue of motion fluctuation\footnote{
consistency, motion fluctuation\footnote{
    Previous works~\cite{gordon2019depth,li2020unsupervised} have alleviated this issue by applying motion smoothness term. This is fair, but only nearby motion vectors are regularized. On the other hands, our regularization method plays with the distribution of motion vectors. Considering the rigidity of the moving objects, \eg, mostly vehicles on traffic roads, we postulate that boosting consistency over a set of whole motion vectors for each object is more helpful to learn semantically plausible object motion field.
% } occurs during this training procedure. 
} occurs during training. 
% This is caused by different photometric error signals of each local pixel.
%As shown in \figref{fluctuation}, the discriminative photometric regions, \eg, headlights, tend to be inferred with high motion response, while the homogeneous regions, \eg, windows, have relatively small motion.
As shown in \figref{fluctuation}, the discriminative regions, \eg, headlights, tend to be inferred with high motion response, while the homogeneous regions, \eg, windows, have relatively small motion.
To mitigate this issue, we propose \emph{contrastive sample consensus (CSAC)} to boost the motion consistency. 

\noindent\textbf{Motion regularization via CSAC}~
Given a semantic prior as a 2D object box and its geometric prior (depth), we design a differentiable regularization module combining traditional random sampling and recent deep learning techniques.
% With a given semantic prior knowledge of 2D object box and its geometric prior knowledge of predicted depth, we design a differentiable regularization module combining traditional random sampling and recent deep learning techniques.
%In order to design the regularization term, first we postulate two assumptions as follows:
This regularization relies on two assumptions:

\begin{assumption} [geometric prior] \label{ass_1}
Each 2D detection box contains a potentially movable object, and it belongs to the foreground region.
\end{assumption}

\begin{assumption} [semantic prior] \label{ass_2}
The motion vectors in each box are distributed into two groups (background: small, object: large), and those belonging to the object group should converge to a single motion vector considering its rigidity under a short time period.
\end{assumption}

\begin{figure}[t]
	\begin{center}
		{\includegraphics[width=0.99\linewidth]{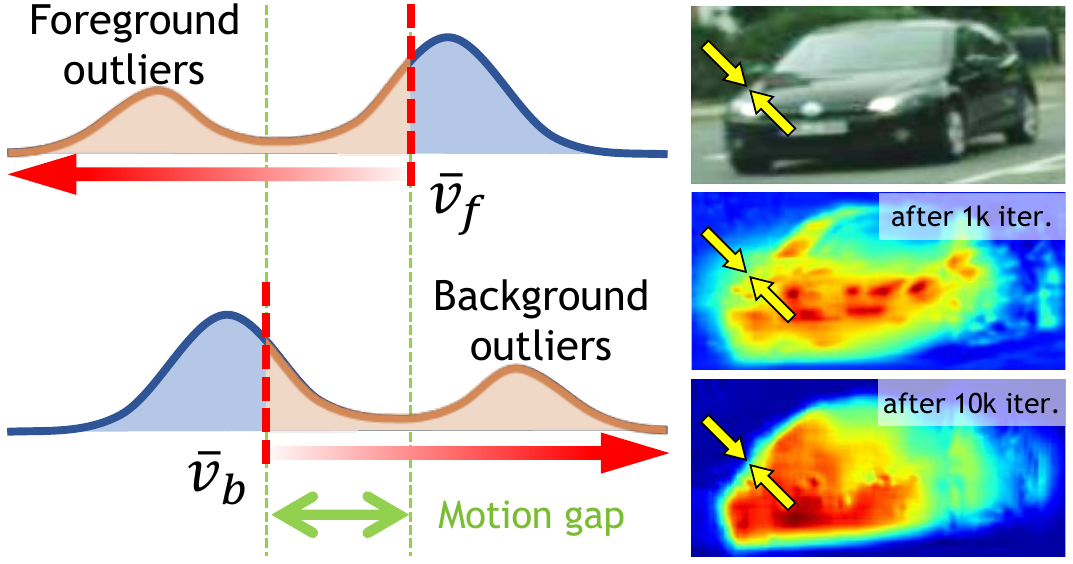}}\hspace{2mm}
	\end{center}
	\vspace{-0mm}
	\caption{
	\textbf{Left}: Schematic of the distributions of foreground and background motions. We penalize small foreground motions and large background motions via CSAC, which eventually increases motion gap between the foreground and background. \textbf{Right}: Visualization of object motion inliers. As the learning progresses ($1k \rightarrow 10k$ iterations), the boundary of motion becomes clearer.
	}
	\label{csac}
	\vspace{-0mm}
\end{figure}

From these assumptions, we train the motions from the foreground and background by \emph{motion-repulsive} embedding as introduced in Algorithm~\ref{alg_1}.
% From those two assumptions, we train motions from foreground and background by \emph{motion-repulsive} embedding as introduced in Algorithm~\ref{alg_1}.
\algnewcommand\Input{\item[\textbf{Input:}]}%
\algnewcommand\Output{\item[\textbf{Output:}]}%
\newcommand{\TextInAlg}[1]{$\langle${#1}$\rangle$}
\begin{algorithm}[t]
\caption{Regularization scheme of residual motion field}
\label{alg_1}
\begin{algorithmic}[1]
\Input{Set of motion vector $\mathbf{V} = \{ v_1, ..., v_n \}$, Set of predicted depth $\mathbf{D} = \{ d_1, ..., d_n \}$ for every $n$-pixel in a detected box}
% \Output{Regularization loss $\calL_{reg}$, Predicted object mask $\mathbf{M}_{obj}$, Regressed object motion $v_{obj}$}
\Output{CSAC loss $\calL_{csac}$ for a detected box}
\Function{Regularizer}{$\mathbf{V}, \mathbf{D}$}
    \State $\calL_{csac} \gets 0$                                             {\footnotesize\Comment{initialize CSAC loss for this detected box}}
    \State $\mathbf{M}_f \gets \Call{FgMask}{\mathbf{D}}$                     {\footnotesize\Comment{$m_k \in \mathbf{M}_f$ is 1 if foreground (\emph{fg})}}
    \State $\mathbf{V}_f \gets \{ v_k | v_k \in \mathbf{V} \land m_k = 1 \}$  {\footnotesize\Comment{\emph{fg} motion set}}
    \State $\mathbf{V}_b \gets \{ v_k | v_k \in \mathbf{V} \land m_k = 0 \}$  {\footnotesize\Comment{\emph{bg} motion set}}
    \For{$\mathcal{V} \gets \{ \mathbf{V}_f, \mathbf{V}_b \}$}                   {\footnotesize\Comment{for both \emph{fg} and \emph{bg} iterations}}
        \State $S_{max} \gets 0$                                                 {\footnotesize\Comment{initialize inlier score}}
        \For{$i \gets 1$ to $N$}                                                 {\footnotesize\Comment{CPU}}
            \State $v_h \gets$ \TextInAlg{random hypothesis from $\mathcal{V}$}
            \State $\mathbf{S} \gets \Call{CalcScore}{v_h, \mathcal{V}}$         {\footnotesize\Comment{for every $v_k \in \mathcal{V}$}}
            \State $S_i \gets \sum \mathbf{S}$
            \If{$S_{max} < S_i$}
                \State $S_{max} \gets S_i$
                \State $\bar{v} \gets \Call{RefineVec}{\mathbf{S}, \mathcal{V}}$ {\footnotesize\Comment{motion refinement}}
            \EndIf
        \EndFor
        % \State $\mathbf{S} \gets \Call{CalcScore}{\bar{v}, \mathcal{V}}$         {\footnotesize\Comment{GPU processing}}
        % \State $\mathbf{M}_{obj} \gets \Call{ObjMask}{\mathbf{S}, \mathbf{M}_f}$
        \State $\calL_{csac} \gets \calL_{csac} + \Call{CalcPenalty}{\bar{v}, \mathcal{V}}$     {\footnotesize\Comment{GPU}}
    \EndFor
    % \State \textbf{return} $\calL_{csac}$, $\mathbf{M}_{obj}$, $v_{obj}$
    \State \textbf{return} $\calL_{csac}$
\EndFunction
\end{algorithmic}
\end{algorithm}
% In this algorithm, first we estimate the initial foreground mask from our predicted depth map on the detection box.
In this algorithm, first we estimate the initial foreground mask from our predicted depth map on the detection box using~\cite{otsu1979threshold} on the depth values (line 3).
% We utilize Otsu's method~\cite{otsu1979threshold} for automatic binary segmentation on the depth map (line 3).
% After splitting foreground and background motion set through the obtained binary mask, we iteratively estimate the representative motion for each set through a random sampling technique.
From this initial binary segmentation mask, we iteratively estimate the representative motion for the foreground and background through a random sampling technique.
During the iteration, we measure the \emph{L1-norm} between the hypothesis $v_h$ and query vectors $v_q$ for each translation axis, and calculate the inlier scores (line 10) as
\begin{equation}
\Scale[0.99]
{
\begin{aligned}
\mathbf{S} = \sum\limits_{v_q \in \mathcal{V}} \mathcal{F}_{inlier} \left( \left| \frac{v_h - v_q}{v_h} \right|_1 \right) ,
\end{aligned}
}
\label{eq_3}
\end{equation}
where $\mathcal{F}_{inlier}$ is designed as
\begin{equation}
\Scale[0.99]
{
\begin{aligned}
\mathcal{F}_{inlier}(\mathbf{x}) = 1 - \sigma ( \alpha \cdot ( \mathbf{x} - \beta ) ) ,
\end{aligned}
}
\label{eq_4}
\end{equation}
which is a soft inlier counting with a \emph{sigmoid} function $\sigma$, proposed by~\cite{brachmann2017dsac,brachmann2018learning}.
In our case, $\alpha$ and $\beta$ are set to 30.0 and 0.2 respectively based on cross-validation.
Then, we measure the iteration score $S_i$ by simply aggregating the inlier scores to find the best hypothesis (line 11). 
The motion refinement is operated by multiplying the query vectors and inlier scores as weights (line 14).
Note that these iterations are processed by the CPU since we do not require gradients for motion estimation.
%Once we get the refined motion vector $\bar{v}$ after iterations, we finally calculate the contrastive penalty loss (line 17), imposed for each foreground and background as
Once we get the refined motion vector $\bar{v}$, we calculate the contrastive penalty loss (line 17), imposed for each foreground and background as
\begin{equation}
\Scale[0.99]
{
\begin{aligned}
\mathcal{L}^{f}_{csac} = \sum\limits_{v_q \in \mathbf{V}_f} \{ 1 - \mathcal{F}_{inlier} ( max ( 0, (|\bar{v}_f| - |v_q|) / |\bar{v}_f| ) ) \} , \\
\mathcal{L}^{b}_{csac} = \sum\limits_{v_q \in \mathbf{V}_b} \{ 1 - \mathcal{F}_{inlier} ( max ( 0, (|v_q| - |\bar{v}_b|) / |\bar{v}_b| ) ) \} ,
% \mathcal{L}^{f}_{csac} = \sum\limits_{v_q \in \mathbf{V}_f} \left\{ 1 - \mathcal{F}_{inlier} \left( max \left( 0, \frac{|\bar{v}| - |v_q|}{|\bar{v}|} \right) \right) \right\} , \\
% \mathcal{L}^{b}_{csac} = \sum\limits_{v_q \in \mathbf{V}_b} \left\{ 1 - \mathcal{F}_{inlier} \left( max \left( 0, \frac{|v_q| - |\bar{v}|}{|\bar{v}|} \right) \right) \right\} ,
\end{aligned}
}
\label{eq_5}
\end{equation}
where $|\bar{v}_f|$ and $|\bar{v}_b|$ are the magnitudes of refined foreground and background motions respectively.
This operation is processed by the GPU to perform the gradient propagation.
In this equation, we penalize the foreground motions smaller than $|\bar{v}_f|$, and background motions larger than $|\bar{v}_b|$, in order to enhance motion contrast between the foreground and background, and this also meets our \textbf{Assumption 2}.
% As represented in \figref{csac}, we show that learning to increase motion contrast guides the networks to learn more accurate motion boundaries.
%\figref{csac} illustrates the importance of our motion contrast enhancement process guiding the network to learn more accurate motion boundaries.
\figref{csac} illustrates the importance of our motion contrast enhancement process to learn more accurate motion boundaries.
Our final residual motion field regularization loss $\calL_{mr}$ is the summation of $\mathcal{L}^{f}_{csac}$ and $\mathcal{L}^{b}_{csac}$ (per-box losses), and normalized to perform per-pixel loss for each mini-batch.

\subsection{Training Scheme}
\label{sec_3_4}

\noindent\textbf{Multi-phase joint training}~
The proposed learning system is composed of complicated submodules.
Although it is possible to gradually converge through end-to-end training, we propose a multi-phase learning technique for efficient and fast convergence.
We summarize the training scheme in \tabref{tab_training}.
Assigned tasks for each phase are jointly trained.
In \emph{phase-1}, since the residual motion branch is inactivated, we set the residual motion to be zero.
% In this subsection, we first introduce our self-supervised losses, and then their multi-phase joint optimization.

\newenvironment{noindlist}
 {\begin{list}{}{\leftmargin=0mm \itemindent=0mm \itemsep=-1mm}}
 {\end{list}}

\noindent\textbf{Self-supervised objective}~
Our complete objective function is composed of \emph{phase-1} loss $\calL^{p_1}$ and \emph{phase-2} loss $\calL^{p_2}$ defined as
\begin{equation}
\Scale[0.99]
{
\begin{aligned}
& \calL^{p_1} = \lambda_{p} \calL_{p} + \lambda_{g} \calL_{g} + \lambda_{s} \calL_{s} + \lambda_{h} \calL_{h}, \\
& \calL^{p_2} = \lambda_{mr} \calL_{mr} + \lambda_{ms} \calL_{ms} + \lambda_{mp} \calL_{mp} + \lambda_{mc} \calL_{mc},
\end{aligned}
}
\label{eq_6}
\end{equation}
where loss weights are grouped as $\Lambda^{p_1} = \{ \lambda_{p}, \lambda_{g}, \lambda_{s}, \lambda_{h} \}$ and $\Lambda^{p_2} = \{ \lambda_{mr}, \lambda_{ms}, \lambda_{mp}, \lambda_{mc} \}$, and each sub-loss is summarized as follows:
\vspace{-0mm}
\begin{noindlist}
\item~$\calL_{p}$ and $\calL_{g}$:~
Photometric and geometric consistency losses defined in \cite{bian2019unsupervised,gao2020attentional}. Occluded and disoccluded regions are masked out by geometric inconsistency map~\cite{lee2021learning}. We replace their global and object-wise motion transformation to our pixel-wise motion representation.
\item~$\calL_{s}$:~
Generic edge-aware depth smoothness term, which is standardized in CC~\cite{ranjan2019competitive}.
\item~$\calL_{h}$:~
Object scale constraint loss with height prior, introduced in Struct2Depth~\cite{casser2019depth}. We use box height as the prior.
\item~$\calL_{mr}$:~
Proposed motion field regularization loss via CSAC.
\item~$\calL_{ms}$:~
We newly propose a reparametrized edge-aware motion smoothness loss. Compared to the \emph{motion-repulsive} embedding by our motion contrastive learning, we add \emph{motion-attractive} embedding to merge the near motion vectors locally. To prevent blurry inference near object boundary, we reparametrize the gradient of the edges with $\tau$ as
\vspace{-0mm}
\begin{equation}
\Scale[0.99]
{
\begin{aligned}
\calL_{ms} = \sum ( \nabla \mathbf{T}^{res} \cdot e^{-\nabla \mathbf{D}/\tau} )^2 ,
\end{aligned}
}
\label{eq_8}
\vspace{-0mm}
\end{equation}
where we set $\tau = 0.1$ in our training. % ffects of different $\tau$ are demonstrated in the \texttt{supplement}.
\item~$\calL_{mp}$ and $\calL_{mc}$:~
Motion sparsity and consistency losses, which are introduced in \cite{gordon2019depth,li2020unsupervised}.
\end{noindlist}

% \item~$\calL_{ma}$:~
% We newly propose an entropy-based motion axis regularization term. This is necessary to prevent an issue of axis bias on the predicted motion, \eg, z-axis motion is excluded when x-axis motion occurs. This is handled by
% \vspace{-3mm}
% \begin{equation}
% \Scale[0.99]
% {
% \begin{aligned}
% \calL_{ma} = \sum\limits_{i \in \{ x, y, z \}} softmax( \mathcal{F}_{inlier} ( max ( 0, (|\bar{v}_f| - |v_q|) / |\bar{v}_f| ) ) ) ,
% \end{aligned}
% }
% \label{eq_7}
% \vspace{-1mm}
% \end{equation}
% where

\begin{table}[t]
\centering
\vspace{-0mm}
\begin{adjustbox}{width=0.99\linewidth}
\setlength{\tabcolsep}{3mm}
\begin{tabular}{cccccc}
    \Xhline{3\arrayrulewidth}
    \multirow{2}[2]{*}{Phase} & \multicolumn{3}{c}{Joint tasks} & \multirow{2}[3]{*}{\shortstack{Training \\ parameters}} & \multirow{2}[3]{*}{Losses} \\
    \cmidrule(l{2pt}r{2pt}){2-4}
    & Depth & Ego- & Res- & & \\
    \Xhline{2\arrayrulewidth}
    $1^{st}$ & \cmark & \cmark & --     & \{$\theta$, $\omega$, $\phi$\}         & $\calL^{p_1}$ \\
    $2^{nd}$ & --     & \cmark & \cmark & \{$\omega$, $\phi$, $\psi$\}           & $\calL^{p_1} + \calL^{p_2}$ \\
    $3^{rd}$ & \cmark & \cmark & \cmark & \{$\theta$, $\omega$, $\phi$, $\psi$\} & $\calL^{p_1} + \calL^{p_2}$ \\
    \Xhline{3\arrayrulewidth}
\end{tabular}
\end{adjustbox}
\vspace{+1mm}
\caption{Multi-phase joint training scheme between the three tasks: depth, ego-motion (ego-), and residual motion (res-) estimation.}
\label{tab_training}
\vspace{-0mm}
\end{table}

% Experiments

\section{Experiments}
\label{sec_4}
In this section, we validate our proposed methods: DAM and CSAC.
We report and analyze experimental results for the tasks of monocular depth and scene flow estimation, motion segmentation, and visual odometry.
%All tasks are performed with a pure monocular-based training, competing methods also rely on the same assumption.
%For fair comparison with the state-of-the-arts, we rule out the impact of different network architectures, \eg, PackNet~\cite{guizilini20203d} and DispNet~\cite{zhou2017unsupervised}.
For the sake of fairness, all competing techniques are purely monocular, moreover, we rule out the impact of different network architectures, \eg, PackNet~\cite{guizilini20203d} and DispNet~\cite{zhou2017unsupervised}.

\subsection{Implementation Details}

\noindent\textbf{Networks}~
We design DepthNet with ResNet18-based ImageNet~\cite{russakovsky2015imagenet} pretrained encoder and decoder structure.
The decoder has the same structure as Monodepth2~\cite{godard2019digging}, and its output is a single-scale inverse depth map with a \emph{sigmoid} activation.
For MotionNet, we use ImageNet pretrained ResNet18 encoder with DAM followed by two motion branches: ego-motion decoder with three convolutional layers, and residual motion field decoder proposed by Gordon \etal~\cite{gordon2019depth}.
Each block of the motion field decoder refines the motion feature from the previous block concatenated with its symmetrically corresponding output feature from the encoding block.

\noindent\textbf{Training}~
Our system is implemented in PyTorch~\cite{paszke2019pytorch} and trained using the ADAM optimizer~\cite{kingma2015adam} with the initial learning rate of $10^{-4}$, $\beta_1 = 0.9$, and $\beta_2 = 0.999$ on $2\times$Nvidia RTX 2080 GPUs.
We set the mini-batch size to 4 and each epoch is trained with $1{,}000$ randomly sampled sequences following the augmentation policy of SC-SfM~\cite{bian2019unsupervised}.
We train \emph{phase-1} and \emph{phase-2} for 10 epochs respectively.
The loss weights, $\Lambda^{p_1}$ and $\Lambda^{p_2}$, are tuned differently depending on the dataset and training phase.
We describe this in detail in the \texttt{supplement}.
To be brief, we empirically found that every objective shows stable convergence when the magnitude of each weighted per-pixel loss ($\lambda \calL$) is 0.05 times the weighted photometric loss ($\lambda_p \calL_p$).
% In addition to the set-up in Table 1, loss weights for the multi-phase training are set to $\Lambda^{p_1} = \{ 3.0, 1.0, 0.1, 0.2 \}$ and $\Lambda^{p_2} = \{ 0.3, 0.5, 1.5, 1.0 \}$.
% About the basis of this loss weighting, we empirically found that every objective shows stable convergence when the magnitude of each weighted per-pixel loss ($\lambda \calL$) is 0.05 times the weighted photometric loss ($\lambda_p \calL_p$).

\noindent\textbf{Dataset}~
Our system is trained and validated in KITTI~\cite{geiger2012we}, Cityscapes~\cite{cordts2016cityscapes}, and Waymo Open Dataset~\cite{sun2020scalability}.
For KITTI and Cityscapes, we utilize the VIS annotations~\cite{lee2021learning} for testing motion segmentation, and detection prior in CSAC training with a random margin up to $10 \%$ for generating detection box. The input resolution is set to $832 \times 256$ for KITTI and Cityscapes, and $480 \times 320$ for Waymo Open Dataset.

\begin{table}[t]
\centering
\vspace{-0mm}
\begin{adjustbox}{width=0.99\linewidth}
\setlength{\tabcolsep}{3mm}
\begin{tabular}{lcccc}
    \Xhline{3\arrayrulewidth}
    Model & \# Params. & AbsRel & SqRel & $\delta_{1.25}$ \\
    \Xhline{2\arrayrulewidth}
    Separated encoders      & 33.45 M & 0.119 & 0.985 & 86.2 \\
    Shared encoder with DAM & \textbf{22.77 M} & \textbf{0.116} & \textbf{0.894} & \textbf{86.9} \\
    \Xhline{3\arrayrulewidth}
\end{tabular}
\end{adjustbox}
\vspace{+1mm}
\caption{\textbf{Ablation study on shared motion encoder with DAM.} Numbers are reported after \emph{phase-3}. Using a single motion encoder yields better performance on monocular depth estimation with fewer parameters.}
\label{tab_ablation1}
\vspace{-0mm}
\end{table}

\newcommand{\mcrot}[4]{\multicolumn{#1}{#2}{\rlap{\rotatebox{#3}{#4}~}}} 
\newcommand*\rot{\rotatebox{90}}

\begin{table}[t]
\centering
\begin{adjustbox}{width=0.88\linewidth}
\setlength{\tabcolsep}{3mm}
\begin{tabular}{ccccccc}
    \Xhline{3\arrayrulewidth}
    \multirow{2}[3]{*}{Models} & \multirow{2}[3]{*}{DAM} & \multirow{2}[3]{*}{CSAC} & \multicolumn{2}{c}{\emph{phase-1}} & \multicolumn{2}{c}{\emph{phase-3}} \\
    \cmidrule(l{2pt}r{2pt}){4-5} \cmidrule(l{2pt}r{2pt}){6-7}
    &  &  & \emph{all} & \emph{obj} & \emph{all} & \emph{obj} \\
    \Xhline{2\arrayrulewidth}
    \emph{A1} & --     & --     & \multirow{2}[0]{*}{0.126} & \multirow{2}[0]{*}{0.202} & 0.120 & 0.199 \\
    \emph{A2} & --     & \cmark &                           &                           & 0.113 & 0.190 \\
    \Xhline{0.01\arrayrulewidth}
    \emph{A3} & \cmark & --     & \multirow{2}[0]{*}{\textbf{0.121}} & \multirow{2}[0]{*}{\textbf{0.196}} & 0.116 & 0.191 \\
    \emph{A4} & \cmark & \cmark &                                    &                                    & \textbf{0.109} & \textbf{0.182} \\
    \Xhline{3\arrayrulewidth}
\end{tabular}
\end{adjustbox}
\vspace{+1mm}
\caption{\textbf{Ablation study on DAM and CSAC.} We measure AbsRel errors after \emph{phase-1} and \emph{phase-3} on both \emph{all} and \emph{obj} areas.}
\label{tab_ablation2}
\end{table}

\subsection{Ablation Study}
% we conduct ablation studies to validate the effect of motion encoding with DAM and motion regularization with CSAC.
To quantify the impact of our motion encoding using DAM and the motion regularization via CSAC, we propose various ablation studies.
In this experiment, the training ($90\%$) and validation ($10\%$) sets are randomly split from KITTI raw monocular videos.
We repeat the training 5 times and average the performance of monocular depth estimation.
First, we perform an ablation to verify our motivation on sharing the motion encoder for MotionNet, while estimating the camera and object motion at the same time.
As shown in \tabref{tab_ablation1}, we achieve better performance with fewer number of trainable parameters, compared to the model with separated encoders.
We, thus, conclude that our invertible attention mechanism enhances the capability of motion disentangling, which produces a better motion feature representation.
Second, we proceed ablation integrated with both DAM and CSAC. 
In this case, we measure AbsRel error on both entire (\emph{all}) and object (\emph{obj}) regions. 
As demonstrated in \tabref{tab_ablation2}, we conduct four ablations (\emph{A1$\sim$A4}) according to our proposed models.
We observe that in \emph{phase-1}, feature extraction with DAM (\emph{A3} and \emph{A4}) has a marginal improvement on depth estimation.
After training MotionNet with residual motion field (\emph{phase-2}), we notice a significant improvement while refining the depth in \emph{phase-3}.
In addition, regularization through CSAC further improves the depth estimation on the object area by providing a rigid body constraint. 
We conclude that our modules play an important role in enhancing the performance of joint depth and motion estimation.
%Therefore, we conclude that our attentive and contrastive motion learning plays an important role in enhancing the performance of joint depth and motion estimation.

\begin{table}[t]
\centering
\vspace{-0mm}
\begin{adjustbox}{width=0.99\linewidth}
\setlength{\tabcolsep}{5pt}
\begin{tabular}{lccccccc}
    \Xhline{3\arrayrulewidth}
    \multirow{2}[3]{*}{Method} & \multirow{2}[3]{*}{\shortstack{Semantic \\ prior}} & \multicolumn{3}{c}{D1} & \multicolumn{3}{c}{D2} \\
    \cmidrule(l{2pt}r{2pt}){3-5} \cmidrule(l{2pt}r{2pt}){6-8}
     &  & \emph{bg} & \emph{fg} & \emph{all} & \emph{bg} & \emph{fg} & \emph{all}  \\
    \Xhline{2\arrayrulewidth}
    DF-Net~\cite{zou2018df}             & -- & --   & --   & 46.5 & -- & -- & 61.5 \\
    GeoNet~\cite{yin2018geonet}         & -- & --   & --   & 49.5 & -- & -- & 58.2 \\
    CC~\cite{ranjan2019competitive}     & -- & 35.0 & 42.7 & 36.2 & -- & -- & -- \\
    SC-SfM~\cite{bian2019unsupervised}  & -- & 36.0 & 46.5 & 37.5 & -- & -- & -- \\
    EPC++ (mono)~\cite{luo2019every}    & -- & 30.7 & 34.4 & 32.7 & \textbf{18.4} & 84.6 & 65.6 \\
    Insta-DM~\cite{lee2021learning}     & instance & \textbf{26.8} & \textbf{30.4} & \textbf{27.4} & \underbar{28.9} & \textbf{32.3} & \textbf{29.4} \\
    \textbf{Ours (DAM+CSAC)}            & box & \underbar{28.6} & \underbar{32.5} & \underbar{29.8} & 30.5 & \underbar{35.7} & \underbar{32.6} \\
    \Xhline{3\arrayrulewidth}
\end{tabular}
\end{adjustbox}
\vspace{+1mm}
\caption{\textbf{Evaluation on the KITTI Scene Flow 2015 training set.} We validate the disparity compared to recent monocular-based training methods. \textbf{Bold}: Best, \underbar{Underbar}: Second best.}
\label{tab_sceneflow}
\vspace{-0mm}
\end{table}

\begin{table}[t]
\centering
\begin{adjustbox}{width=0.92\linewidth}
\setlength{\tabcolsep}{3mm}
\begin{tabular}{ccccc}
    \Xhline{2\arrayrulewidth}
     & Before reg. & After reg. & Li \etal~\cite{li2020unsupervised} & CC~\cite{ranjan2019competitive} \\
    \Xhline{1\arrayrulewidth}
    KITTI-VIS      & 0.483 & \textbf{0.813} & 0.689 & 0.571 \\
    Cityscapes-VIS & 0.416 & \textbf{0.785} & 0.620 & --    \\
    \Xhline{2\arrayrulewidth}
\end{tabular}
\end{adjustbox}
\vspace{+1mm}
\caption{\textbf{Results (mean IoU) of object (vehicle class) segmentation with a given box prior.} The results show that CSAC regularization improve the performance on semantic perception task.}
\label{tab_segmentation}
\end{table}

\begin{figure*}[t] 
	\centering
    \includegraphics[width=0.99\linewidth]{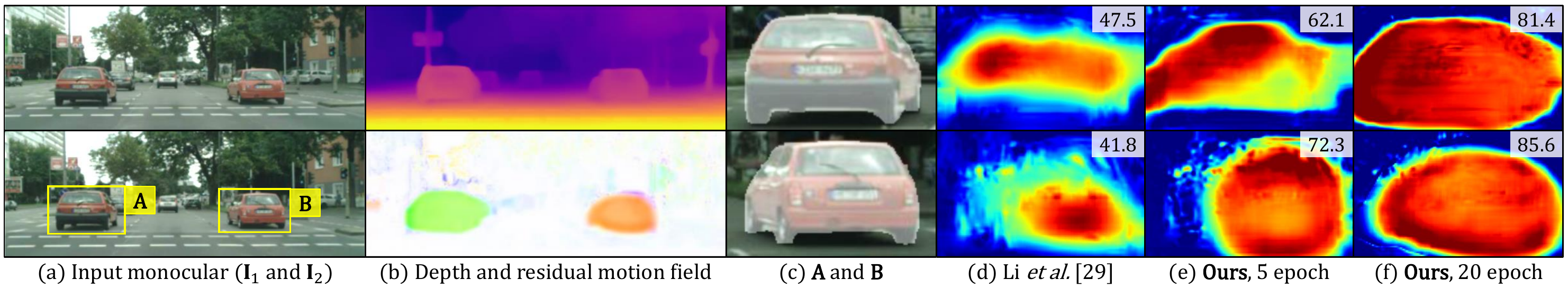}
	\caption{\textbf{Qualitative results of our depth and residual motion field on Cityscapes test set.} Each scene shows (a) consecutive input images with a box prior, (b) our networks outputs, (c) object images with GT mask, (d) motion inliers of the previous method~\cite{li2020unsupervised} and ours after training (e) 5 and (f) 20 epoch. The inlier maps are normalized in the same scale, and we indicate their mean IoU from the GT.
	}
	\label{qualitative}
\end{figure*}

\begin{table*}[t]
\begin{minipage}{.67\linewidth}
    \centering
    \vspace{-0mm}
    \begin{adjustbox}{width=0.99\textwidth}
    \setlength{\tabcolsep}{2mm}
    \begin{tabular}{lccccccccccc}

    \Xhline{4\arrayrulewidth}
    \multirow{2}{*}{Method} & \multirow{2}{*}{\shortstack{Semantic \\ prior}} & \multirow{2}{*}{Training} & \multirow{2}{*}{Test} & \multicolumn{4}{c}{Error metric $\downarrow$} & & \multicolumn{3}{c}{Accuracy metric $\uparrow$}  \\
    \cline{5-8} \cline{10-12}
     &  &  &  & AbsRel  & SqRel & RMSE & RMSE log & & $\delta < 1.25$ & $\delta < 1.25^2$ & $\delta < 1.25^3$ \\ \Xhline{2\arrayrulewidth}
    
    Monodepth2~\cite{godard2019digging}         & -- & K     & K & 0.115 & 0.903 & 4.863 & \underbar{0.193} &  & \underbar{0.877} & \underbar{0.959} & \textbf{0.981} \\
    Li~\etal~\cite{li2020unsupervised}          & -- & K     & K & 0.130 & 0.950 & 5.138 & 0.209 &  & 0.843 & 0.948 & 0.978 \\
    Struct2Depth~\cite{casser2019depth}         & instance & K & K & 0.141 & 1.026 & 5.290 & 0.215 &  & 0.816 & 0.945 & \underbar{0.979} \\
    Gordon~\etal~\cite{gordon2019depth}         & box & K & K & 0.128 & 0.959 & 5.230 & 0.212 &  & 0.845 & 0.947 & 0.976 \\
    SGDepth~\cite{klingner2020selfsupervised}   & segment & K & K & \textbf{0.113} & \textbf{0.835} & \textbf{4.693} & \textbf{0.191} &  & \textbf{0.879} & \textbf{0.961} & \textbf{0.981} \\
    \textbf{Ours (DAM+CSAC)}                    & box & K & K & \underbar{0.114} & \underbar{0.876} & \underbar{4.715} & \textbf{0.191} &  & 0.872 & 0.955 & \textbf{0.981} \\
    \Xhline{1\arrayrulewidth}

    Gordon~\etal~\cite{gordon2019depth}         & box & C+K & K & \underbar{0.124} & \underbar{0.930} & \underbar{5.120} & \underbar{0.206} &  & \underbar{0.851} & \underbar{0.950} & \underbar{0.978} \\
    \textbf{Ours (DAM+CSAC)}                    & box & C+K & K & \textbf{0.111} & \textbf{0.805} & \textbf{4.708} & \textbf{0.187} &  & \textbf{0.875} & \textbf{0.962} & \textbf{0.981} \\
    \Xhline{1\arrayrulewidth}
    
    Li~\etal~\cite{li2020unsupervised}          & -- & C     & C & \underbar{0.119} & \underbar{1.290} & 6.980 & \underbar{0.190} &  & \underbar{0.846} & \textbf{0.952} & \textbf{0.982} \\
    Struct2Depth~\cite{casser2019unsupervised}  & instance & C & C & 0.145 & 1.737 & 7.280 & 0.205 &  & 0.813 & 0.942 & 0.978 \\
    Gordon~\etal~\cite{gordon2019depth}         & box & C & C & 0.127 & 1.330 & \underbar{6.960} & 0.195 &  & 0.830 & 0.947 & \underbar{0.981} \\
    \textbf{Ours (DAM+CSAC)}                    & box & C & C & \textbf{0.116} & \textbf{1.213} & \textbf{6.695} & \textbf{0.186} &  & \textbf{0.852} & \underbar{0.951} & \textbf{0.982} \\
    \Xhline{1\arrayrulewidth}
    
    Monodepth2~\cite{godard2019digging}         & -- & W & W & 0.168 & 1.738 & 7.947 & 0.230 &  & -- & -- & -- \\
    Li~\etal~\cite{li2020unsupervised}          & -- & W & W & \underbar{0.162} & \underbar{1.711} & \underbar{7.833} & \underbar{0.223} &  & -- & -- & -- \\
    Struct2Depth~\cite{casser2019depth}         & instance & W & W & 0.180 & 1.782 & 8.583 & 0.244 &  & -- & -- & -- \\
    \textbf{Ours (DAM+CSAC)}                    & box & W & W & \textbf{0.148} & \textbf{1.686} & \textbf{7.420} & \textbf{0.210} &  & -- & -- & -- \\
    \Xhline{4\arrayrulewidth}

    \end{tabular}
    \end{adjustbox}
    \vspace{+1mm}
    \caption{Monocular depth estimation results on the KITTI (K) Eigen test set, Cityscapes (C) test set, and Waymo Open Dataset (W). Models pretrained on Cityscapes and fine-tuned on KITTI are denoted by `C+K'. Due to the page limit, we only indicate our final model (DAM+CSAC), and methods using strong semantic priors (\eg, video instance segmentation) are ruled out. Full table is demonstrated in the \texttt{supplement}. For each partition, \textbf{Bold}: Best, \underbar{Underbar}: Second best.}
    \label{tab_kitti}

\end{minipage}
\hspace{+1mm}
\begin{minipage}{.31\linewidth}

    \centering
    \renewcommand{\tabcolsep}{1.5mm}
    \begin{adjustbox}{width=0.99\textwidth}
    \begin{tabular}{lcc}
    \Xhline{3\arrayrulewidth}
    Method & Seq. 09 & Seq. 10 \\
    \Xhline{1\arrayrulewidth}
    SfM-Learner~\cite{zhou2017unsupervised}     & $0.021\pm0.017$ & $0.020\pm0.015$ \\
    GeoNet~\cite{yin2018geonet}                 & $0.012\pm0.007$ & $0.012\pm0.009$ \\
    CC~\cite{ranjan2019competitive}             & $0.012\pm0.007$ & $0.012\pm0.008$ \\
    Struct2Depth~\cite{casser2019depth}         & $0.011\pm0.006$ & $0.011\pm0.010$ \\
    GLNet~\cite{chen2019self}                   & $0.011\pm0.006$ & $\mathbf{0.011\pm0.009}$ \\
    SGDepth~\cite{klingner2020selfsupervised}   & $0.017\pm0.009$ & $0.014\pm0.010$ \\
    \Xhline{1\arrayrulewidth}
    \textbf{Ours (w/o DAM)}                              & $0.012\pm0.006$ & $\mathbf{0.011\pm0.009}$ \\
    \textbf{Ours (w/ DAM)}                               & $\mathbf{0.010\pm0.011}$ & $\mathbf{0.011\pm0.009}$ \\
    \Xhline{3\arrayrulewidth}
    \end{tabular}
    \end{adjustbox}
    \vspace{+0.5mm}
    \caption{Absolute trajectory error (ATE) on KITTI-VO.}
    \label{tab_odom}

    \vspace{+2mm}
    \centering
    \begin{adjustbox}{width=0.99\textwidth}
    \renewcommand{\tabcolsep}{2.5mm}
    \begin{tabular}{lcccc}
    \Xhline{3\arrayrulewidth}
    \multirow{2}[3]{*}{Method} & \multicolumn{2}{c}{Seq. 09} & \multicolumn{2}{c}{Seq. 10} \\
    \cmidrule(l{2pt}r{2pt}){2-3} \cmidrule(l{2pt}r{2pt}){4-5}
    & $t_{err}$ & $r_{err}$ & $t_{err}$ & $r_{err}$  \\
    \Xhline{2\arrayrulewidth}
    GeoNet~\cite{yin2018geonet} & 39.4 & 14.3 & 29.0 & 8.6 \\
    SC-SfM~\cite{bian2019unsupervised} & 11.2 & 3.4 & 10.1 & 5.0 \\
    \Xhline{1\arrayrulewidth}
    \textbf{Ours (w/o DAM)} & 9.7 & 3.4 & 9.9 & 4.8 \\
    \textbf{Ours (w/ DAM)}  & \textbf{8.9} & \textbf{3.3} & \textbf{9.5} & \textbf{4.7} \\
    \Xhline{3\arrayrulewidth}
    \end{tabular}
    \end{adjustbox}
    \vspace{+1mm}
    \caption{Relative translation $t_{err}$ ($\%$) and rotation $r_{err}$ ($\sfrac{^{\circ}}{100m}$) errors on KITTI-VO.}
	\label{tab_odom2}

\end{minipage}
\vspace{-0mm}
\end{table*}

\subsection{Monocular Scene Flow Estimation}

%in order to validate our pixel-wise motion and depth estimation simultaneously, we demonstrate the results of monocular scene flow estimation on the KITTI Scene Flow 2015 training set, as shown in \tabref{tab_sceneflow}.
To validate our pixel-wise motion and depth estimation simultaneously, we assessed our monocular scene flow estimation on the KITTI Scene Flow 2015 training set, as shown in \tabref{tab_sceneflow}.
We compare our method with existing monocular-based training methods.
Compared to the methods not using semantic priors, we achieve more than $57.8 \%$ accuracy gain for estimating the disparity of objects on the target image (D2-\emph{fg}).
Despite using weaker priors, we obtain competitive results against techniques relying on strong semantic prior, such as, Insta-DM~\cite{lee2021learning}.
%Although we may not surpass all previous methods, considering that Insta-DM~\cite{lee2021learning} is trained with a strong semantic prior of video instance segmentation, we show that our method is comparable.

\subsection{Motion Segmentation}

We demonstrate object motion segmentation on KITTI Scene Flow 2015 training set and Cityscapes with VIS annotation~\cite{lee2021learning}.
This is enabled by leveraging inlier scores (threshold of 0.5) of the best hypothesis, as designed in \eqnref{eq_4}. 
\tabref{tab_segmentation} demonstrates that our regularization with contrastive learning based on the geometric prior significantly improves the performance of semantic perception task. 
We demonstrate qualitative results in \figref{qualitative}.

\subsection{Monocular Depth Estimation and Visual Odometry}

Finally, we provide comparisons to state-of-the-arts~\cite{godard2019digging,li2020unsupervised,casser2019depth,gordon2019depth,klingner2020selfsupervised} of self-supervised monocular depth and ego-motion estimation based on monocular training.
We compare the depth estimation on KITTI Eigen split~\cite{eigen2014depth}, Cityscapes~\cite{cordts2016cityscapes}, and Waymo Open Dataset~\cite{sun2020scalability}, and all the compared methods are based on the ResNet18 backbone.
As shown in \tabref{tab_kitti}, our final model with DAM and CSAC outperforms all published self-supervised methods with weak semantic priors (up to box prior). 
Qualitative results are demonstrated in the \texttt{supplement}.

We also demonstrate visual odometry on KITTI-VO in \tabref{tab_odom} and \tabref{tab_odom2}.
In these experiments, our model with DAM outperforms state-of-the-arts using monocular self-supervised training.
%In these experiments, we focus on the role of DAM, and our model with DAM outperforms state-of-the-arts using monocular self-supervised training.
We conclude that our attention module favorably works in estimating the camera ego-motion.

\section{Conclusion}

We proposed a novel self-supervised learning framework to estimate the motion field of a dynamic scene from a monocular camera.
First, our approach heavily relies on a novel attention module dedicated to the disentanglement of the camera ego-motion and the objects' motions, which has proven to be effective to improve the overall performance of our network.
Second, we designed an object motion field estimation through contrastive sample consensus.
With given geometric and semantic priors, we leverage a motion-repulsive embedding near object motion boundaries to estimate more accurate motion field.
The effectiveness of our system has been demonstrated on various driving datasets.
%One of the limitations of the proposed approach is the rigid body assumption of the moving objects. 
One remaining limitation is the rigid body assumption of the moving objects.
While our approach is suitable for a traffic scene containing vehicles, it is not appropriate for deformable objects such as pedestrians. 
Therefore, we leave this problem as our future direction to improve the applicability of the technique to more diverse scenarios.

\section*{Acknowledgement}

This work was supported under the framework of international cooperation program managed by the National Research Foundation of Korea (NRF-2020M3H8A1115028, FY2021).
This work was supported in part by the Institute of Information and Communications Technology Planning and Evaluation (IITP) Grant funded by the Korea Government (MSIT) (Artificial Intelligence Innovation Hub) under Grant 2021-0-02068.

{\small
\bibliographystyle{ieee_fullname}
\bibliography{__egbib}
}

\clearpage

\renewcommand{\thesection}{A\arabic{section}}
\renewcommand{\thetable}{A\arabic{table}}
\renewcommand{\thefigure}{A\arabic{figure}}
\renewcommand{\theequation}{A\arabic{equation}}

\setcounter{section}{0} 
\setcounter{figure}{0} 
\setcounter{table}{0} 

% \begin{appendices}

%%%%%%%%% TITLE
% \title{Supplementary Material: \linebreak Attentive and Contrastive Learning for Joint Depth and Motion Field Estimation}

% \author{Seokju Lee \hspace{5 mm}
%         Francois Rameau \hspace{5 mm}
%         Fei Pan \hspace{5 mm}
%         In So Kweon\\
        
%         Korea Advanced Institute of Science and Technology (KAIST)\\
    
%         {\tt\small \{seokju91,rameau.fr,feipan664\}@gmail.com, iskweon77@kaist.ac.kr}
%         % For a paper whose authors are all at the same institution,
%         % omit the following lines up until the closing ``}''.
%         % Additional authors and addresses can be added with ``\and'',
%         % just like the second author.
%         % To save space, use either the email address or home page, not both

% }

% \maketitle

\section*{\fontsize{16}{26}\selectfont Supplementary Material}
% \section*{\fontsize{16}{26}\selectfont Appendix}

In this supplementary material, we present additional details on the models, training scheme, and experiments that could not be included in the main text due to space constraints. All tables, figures, equations, and references in this supplementary file are self-contained.

% \noindent\textbf{Source codes}~
% We attach the source codes (based on the PyTorch library) for the implementation of the proposed networks and CSAC with the \texttt{supplement}.

\noindent\textbf{Contents of the supplementary material}~
The supplementary material is composed of the following. (1) Details of our models. (2) Additional ablation study such as networks design and hyper-parameter settings. (3) Additional experimental results.

% \noindent\textbf{Typo in the main paper}~
% There is a minor typo in Fig 3. of the main submission.
% We will change ($\mathbf{M}_c$, $\mathbf{M}_s$) to ($\mathbf{A}_c$, $\mathbf{A}_s$) for the consistency with L413 and L416 of our main paper.

\begin{table*}[t]
    \vspace{-0mm}
	\centering
	\renewcommand{\tabcolsep}{3mm}
	\begin{adjustbox}{width=0.85\textwidth}
		\begin{tabular}{cgcgcgcg}   % {cccccccc}   % {cgcgcgcg}
			\toprule[0.7mm]
			Type & Name & Kernel & Stride & Channel I/O & In. resol. & Out. resol. & Input \\ 
            \midrule[0.35mm]
            \multirow{10}{*}{Encoder} 
			& conv\_0    & $7\times 7$ & 2 & 3/64   & $256\times 832$ & $128\times 416$ & image \\
			\cmidrule[0.1mm](l{2pt}r{2pt}){2-8}
			& maxpool\_1  & $3\times 3$ & 2 & 64/64  & $128\times 416$ & $64\times 208$  & conv\_0 \\
			& ResBlock\_1\_1 & -- & 1 & 64/64  & $64\times 208$ & $64\times 208$ & maxpool\_1 \\
			& ResBlock\_1\_2 & -- & 1 & 64/64  & $64\times 208$ & $64\times 208$ & ResBlock\_1\_1 \\
			\cmidrule[0.1mm](l{2pt}r{2pt}){2-8}
			& ResBlock\_2\_1 & -- & 2 & 64/128  & $64\times 208$ & $32\times 104$ & ResBlock\_1\_2 \\
			& ResBlock\_2\_2 & -- & 1 & 128/128 & $32\times 104$ & $32\times 104$ & ResBlock\_2\_1 \\
			\cmidrule[0.1mm](l{2pt}r{2pt}){2-8}
			& ResBlock\_3\_1 & -- & 2 & 128/256 & $32\times 104$ & $16\times 52$ & ResBlock\_2\_2 \\
			& ResBlock\_3\_2 & -- & 1 & 256/256 & $16\times 52$  & $16\times 52$ & ResBlock\_3\_1 \\
			\cmidrule[0.1mm](l{2pt}r{2pt}){2-8}
			& ResBlock\_4\_1 & -- & 2 & 256/512 & $16\times 52$ & $8\times 26$ & ResBlock\_3\_2 \\
			& ResBlock\_4\_2 & -- & 1 & 512/512 & $8\times 26$  & $8\times 26$ & ResBlock\_4\_1 \\
			
			\midrule[0.35mm]
			\multirow{21}{*}{Decoder}
			& conv\_1\_1 & $3\times 3$ & 1 & 512/256 & $8\times 26$ & $8\times 26$   & ResBlock\_4\_2 \\
			& upsample\_1  & -- & --       & 256/256 & $8\times 26$  & $16\times 52$ & conv\_1\_1 \\
			& concat\_1    & -- & --       & (256,256)/512 & $16\times 52$ & $16\times 52$ & upsample\_1, ResBlock\_3\_2 \\
			& conv\_1\_2 & $3\times 3$ & 1 & 512/256 & $16\times 52$ & $16\times 52$ & concat\_1 \\
			\cmidrule[0.1mm](l{2pt}r{2pt}){2-8}
			& conv\_2\_1 & $3\times 3$ & 1 & 256/128 & $16\times 52$  & $16\times 52$  & conv\_1\_2 \\
			& upsample\_2  & -- & --       & 128/128 & $16\times 52$  & $32\times 104$ & conv\_2\_1 \\
			& concat\_2    & -- & --       & (128,128)/256 & $32\times 104$ & $32\times 104$ & upsample\_2, ResBlock\_2\_2 \\
			& conv\_2\_2 & $3\times 3$ & 1 & 256/128 & $32\times 104$ & $32\times 104$ & concat\_2 \\
			\cmidrule[0.1mm](l{2pt}r{2pt}){2-8}
			& conv\_3\_1 & $3\times 3$ & 1 & 128/64 & $32\times 104$ & $32\times 104$ & conv\_2\_2 \\
			& upsample\_3  & -- & --       & 64/64  & $32\times 104$ & $64\times 208$ & conv\_3\_1 \\
			& concat\_3    & -- & --       & (64,64)/128 & $64\times 208$ & $64\times 208$ & upsample\_3, ResBlock\_1\_2 \\
			& conv\_3\_2 & $3\times 3$ & 1 & 128/64 & $64\times 208$ & $64\times 208$ & concat\_3 \\
			\cmidrule[0.1mm](l{2pt}r{2pt}){2-8}
			& conv\_4\_1 & $3\times 3$ & 1 & 64/32 & $64\times 208$  & $64\times 208$  & conv\_3\_2 \\
			& upsample\_4  & -- & --       & 32/32 & $64\times 208$  & $128\times 416$ & conv\_4\_1 \\
			& concat\_4    & -- & --       & (32,64)/96 & $128\times 416$ & $128\times 416$ & upsample\_4, conv\_0 \\
			& conv\_4\_2 & $3\times 3$ & 1 & 96/32 & $128\times 416$ & $128\times 416$ & concat\_4 \\
			\cmidrule[0.1mm](l{2pt}r{2pt}){2-8}
			& conv\_5\_1 & $3\times 3$ & 1 & 32/16 & $128\times 416$ & $128\times 416$ & conv\_4\_2 \\
			& upsample\_5  & -- & --       & 16/16 & $128\times 416$ & $256\times 832$ & conv\_5\_1 \\
			& conv\_5\_2 & $3\times 3$ & 1 & 16/16 & $256\times 832$ & $256\times 832$ & upsample\_5 \\
			& dispconv     & $3\times 3$ & 1 & 16/1  & $256\times 832$ & $256\times 832$ & conv\_5\_2 \\
			& sigmoid      & -- & --         & 1/1   & $256\times 832$ & $256\times 832$ & dispconv \\
			
			\bottomrule[0.7mm]
		\end{tabular}
	\end{adjustbox}
	\vspace{+2mm}
	\caption{Details of DepthNet.}
	\label{tab_depthnet}
	\vspace{-0mm}
\end{table*}

\begin{table*}[t]
    \vspace{+5mm}
	\centering
	\renewcommand{\tabcolsep}{3mm}
	\begin{adjustbox}{width=0.85\textwidth}
		\begin{tabular}{cgcgcgcg}   % {cccccccc}   % {cgcgcgcg}
			\toprule[0.6mm]
			Type & Name & Kernel & Stride & Channel I/O & In. resol. & Out. resol. & Input \\ 
            \midrule[0.3mm]
            \multirow{3}{*}{\shortstack{Spatial \\ attention}} 
			& conv\_1\_1     & $ 1 \times 1 $ & 1 & c/$\frac{c}{2}$  & $ h \times w $ & $ h \times w $ & feature \\
			& conv\_1\_2     & $ 1 \times 1 $ & 1 & $\frac{c}{2}$/1  & $ h \times w $ & $ h \times w $ & conv\_1\_1 \\
			& softmax        & -- & -- & 1/1 & $ h \times w $ & $ h \times w $ & conv\_1\_2 \\
			\midrule[0.3mm]
			\multirow{4}{*}{Transform}
			& matmul     & -- & -- & c/c & $ h \times w $ & $ 1 \times 1 $ & feature, softmax \\
			& conv\_2\_1 & $ 1 \times 1 $ & 1 & c/$\frac{c}{4}$ & $ 1 \times 1 $ & $ 1 \times 1 $ & matmul \\
			& conv\_2\_2 & $ 1 \times 1 $ & 1 & $\frac{c}{4}$/c & $ 1 \times 1 $ & $ 1 \times 1 $ & conv\_2\_1 \\
			& sum        & -- & --            & c/c & $ h \times w $ & $ h \times w $ & feature, conv\_2\_2 \\

			\bottomrule[0.6mm]
		\end{tabular}
	\end{adjustbox}
	\vspace{+2mm}
	\caption{Details of DAM (single attention module).}
	\label{tab_dam}
	\vspace{-0mm}
\end{table*}

\begin{table*}[t]
    \vspace{-0mm}
	\centering
	\renewcommand{\tabcolsep}{3mm}
	\begin{adjustbox}{width=0.85\textwidth}
		\begin{tabular}{cgcgcgcg}   % {cccccccc}   % {cgcgcgcg}
			\toprule[0.8mm]
			Type & Name & Kernel & Stride & Channel I/O & In. resol. & Out. resol. & Input \\ 
            \midrule[0.4mm]
            \multirow{13}{*}{Encoder} 
			& conv\_0 & $7\times 7$ & 2 & (3,3,1,1)/64  & $256\times 832$ & $128\times 416$ & images, depth maps \\
% 			& DAM\_0   & --         & 1 & 64/64 & $128\times 416$ & $128\times 416$ & conv\_0 \\
			
			\cmidrule[0.1mm](l{2pt}r{2pt}){2-8}
			& maxpool\_1     & $3\times 3$ & 2 & 64/64 & $128\times 416$ & $64\times 208$ & conv\_0 \\
			& ResBlock\_1\_1 & --          & 1 & 64/64 & $64\times 208$  & $64\times 208$ & maxpool\_1 \\
			& ResBlock\_1\_2 & --          & 1 & 64/64 & $64\times 208$  & $64\times 208$ & ResBlock\_1\_1 \\
% 			& DAM\_1         & --          & 1 & 64/64 & $64\times 208$  & $64\times 208$ & ResBlock\_1\_2 \\

			\cmidrule[0.1mm](l{2pt}r{2pt}){2-8}
			& ResBlock\_2\_1 & -- & 2 & 64/128  & $64\times 208$ & $32\times 104$ & ResBlock\_1\_2 \\
			& ResBlock\_2\_2 & -- & 1 & 128/128 & $32\times 104$ & $32\times 104$ & ResBlock\_2\_1 \\
			& DAM\_2         & -- & 1 & 128/128 & $32\times 104$ & $32\times 104$ & ResBlock\_2\_2 \\
			
			\cmidrule[0.1mm](l{2pt}r{2pt}){2-8}
			& ResBlock\_3\_1 & -- & 2 & 128/256 & $32\times 104$ & $16\times 52$ & DAM\_2 \\
			& ResBlock\_3\_2 & -- & 1 & 256/256 & $16\times 52$  & $16\times 52$ & ResBlock\_3\_1 \\
			& DAM\_3         & -- & 1 & 256/256 & $16\times 52$  & $16\times 52$ & ResBlock\_3\_2 \\
			
			\cmidrule[0.1mm](l{2pt}r{2pt}){2-8}
			& ResBlock\_4\_1 & -- & 2 & 256/512 & $16\times 52$ & $8\times 26$ & DAM\_3 \\
			& ResBlock\_4\_2 & -- & 1 & 512/512 & $8\times 26$  & $8\times 26$ & ResBlock\_4\_1 \\
			& DAM\_4         & -- & 1 & 512/512 & $8\times 26$  & $8\times 26$ & ResBlock\_4\_2 \\
			
			\midrule[0.4mm]
			\multirow{5}{*}{\shortstack{Ego-motion \\ decoder}}
			& conv\_1 & $1\times 1$ & 1 & 512/256 & $8\times 26$ & $8\times 26$ & DAM\_4 \\
			& conv\_2 & $3\times 3$ & 1 & 256/256 & $8\times 26$ & $8\times 26$ & conv\_1 \\
			& conv\_3 & $3\times 3$ & 1 & 256/256 & $8\times 26$ & $8\times 26$ & conv\_2 \\
			& conv\_4 & $1\times 1$ & 1 & 256/6   & $8\times 26$ & $8\times 26$ & conv\_3 \\
			& avgpool & -- & --         & 6/6     & $8\times 26$ & $1\times 1$  & conv\_4 \\

			\midrule[0.4mm]
			\multirow{42}{*}{\shortstack{Res-motion \\ decoder}}
			& mofconv\_0   & $1\times 1$ & 1 & 512/3    & $8\times 26$ & $8\times 26$ & DAM\_4 \\
			\cmidrule[0.1mm](l{2pt}r{2pt}){2-8}

			& concat\_1\_1 & -- & --         & (3,512)/515 & $8\times 26$ & $8\times 26$ & mofconv\_0, DAM\_4 \\
			& conv\_1\_a   & $3\times 3$ & 1 & 515/512  & $8\times 26$ & $8\times 26$ & concat\_1\_1 \\
			& conv\_1\_b   & $3\times 3$ & 1 & 515/512  & $8\times 26$ & $8\times 26$ & concat\_1\_1 \\
			& concat\_1\_2 & -- & --         & (512,512)/1024 & $8\times 26$ & $8\times 26$ & conv\_1\_a, conv\_1\_b \\
			& mofconv\_1   & $1\times 1$ & 1 & 1024/3   & $8\times 26$ & $8\times 26$ & concat\_1\_2 \\
			& sum\_1       & -- & --         & (3,3)/3      & $8\times 26$ & $8\times 26$ & mofconv\_0, mofconv\_1 \\
			\cmidrule[0.1mm](l{2pt}r{2pt}){2-8}
			
			& upsample\_2  & -- & --         & 3/3     & $8\times 26$  & $16\times 52$ & sum\_1 \\
			& concat\_2\_1 & -- & --         & (3,256)/259   & $16\times 52$ & $16\times 52$ & upsample\_2, DAM\_3 \\
			& conv\_2\_a   & $3\times 3$ & 1 & 259/256 & $16\times 52$ & $16\times 52$ & concat\_2\_1 \\
			& conv\_2\_b   & $3\times 3$ & 1 & 259/256 & $16\times 52$ & $16\times 52$ & concat\_2\_1 \\
			& concat\_2\_2 & -- & --         & (256,256)/512 & $16\times 52$ & $16\times 52$ & conv\_2\_a, conv\_2\_b \\
			& mofconv\_2   & $1\times 1$ & 1 & 512/3   & $16\times 52$ & $16\times 52$ & concat\_2\_2 \\
			& sum\_2       & -- & --         & (3,3)/3     & $16\times 52$ & $16\times 52$ & mofconv\_2, upsample\_2 \\
			\cmidrule[0.1mm](l{2pt}r{2pt}){2-8}
			
			& upsample\_3  & -- & --         & 3/3     & $16\times 52$  & $32\times 104$ & sum\_2 \\
			& concat\_3\_1 & -- & --         & (3,128)/131   & $32\times 104$ & $32\times 104$ & upsample\_3, DAM\_2 \\
			& conv\_3\_a   & $3\times 3$ & 1 & 131/128 & $32\times 104$ & $32\times 104$ & concat\_3\_1 \\
			& conv\_3\_b   & $3\times 3$ & 1 & 131/128 & $32\times 104$ & $32\times 104$ & concat\_3\_1 \\
			& concat\_3\_2 & -- & --         & (128,128)/256 & $32\times 104$ & $32\times 104$ & conv\_3\_a, conv\_3\_b \\
			& mofconv\_3   & $1\times 1$ & 1 & 256/3   & $32\times 104$ & $32\times 104$ & concat\_3\_2 \\
			& sum\_3       & -- & --         & (3,3)/3     & $32\times 104$ & $32\times 104$ & mofconv\_3, upsample\_3 \\
			\cmidrule[0.1mm](l{2pt}r{2pt}){2-8}
			
			& upsample\_4  & -- & --         & 3/3    & $32\times 104$ & $64\times 208$ & sum\_3 \\
			& concat\_4\_1 & -- & --         & (3,64)/67   & $64\times 208$ & $64\times 208$ & upsample\_4, ResBlock\_1\_2 \\
			& conv\_4\_a   & $3\times 3$ & 1 & 67/64  & $64\times 208$ & $64\times 208$ & concat\_4\_1 \\
			& conv\_4\_b   & $3\times 3$ & 1 & 67/64  & $64\times 208$ & $64\times 208$ & concat\_4\_1 \\
			& concat\_4\_2 & -- & --         & (64,64)/128 & $64\times 208$ & $64\times 208$ & conv\_4\_a, conv\_4\_b \\
			& mofconv\_4   & $1\times 1$ & 1 & 128/3  & $64\times 208$ & $64\times 208$ & concat\_4\_2 \\
			& sum\_4       & -- & --         & (3,3)/3    & $64\times 208$ & $64\times 208$ & mofconv\_4, upsample\_4 \\
			\cmidrule[0.1mm](l{2pt}r{2pt}){2-8}
			
			& upsample\_5  & -- & --         & 3/3    & $64\times 208$ & $128\times 416$ & sum\_4 \\
			& concat\_5\_1 & -- & --         & 3/67   & $128\times 416$ & $128\times 416$ & upsample\_5, conv\_0 \\
			& conv\_5\_a   & $3\times 3$ & 1 & 67/64  & $128\times 416$ & $128\times 416$ & concat\_5\_1 \\
			& conv\_5\_b   & $3\times 3$ & 1 & 67/64  & $128\times 416$ & $128\times 416$ & concat\_5\_1 \\
			& concat\_5\_2 & -- & --         & 64/128 & $128\times 416$ & $128\times 416$ & conv\_5\_a, conv\_5\_b \\
			& mofconv\_5   & $1\times 1$ & 1 & 128/3  & $128\times 416$ & $128\times 416$ & concat\_5\_2 \\
			& sum\_5       & -- & --         & (3,3)/3    & $128\times 416$ & $128\times 416$ & mofconv\_5, upsample\_5 \\
			\cmidrule[0.1mm](l{2pt}r{2pt}){2-8}
			
			& upsample\_6  & -- & --         & 3/3  & $128\times 416$ & $256\times 832$ & sum\_4 \\
			& concat\_6\_1 & -- & --         & (3,3,3,1,1)/11 & $256\times 832$ & $256\times 832$ & upsample\_6, images, depth maps \\
			& conv\_6\_a   & $3\times 3$ & 1 & 11/8 & $256\times 832$ & $256\times 832$ & concat\_6\_1 \\
			& conv\_6\_b   & $3\times 3$ & 1 & 11/8 & $256\times 832$ & $256\times 832$ & concat\_6\_1 \\
			& concat\_6\_2 & -- & --         & (8,8)/16 & $256\times 832$ & $256\times 832$ & conv\_6\_a, conv\_6\_b \\
			& mofconv\_6   & $1\times 1$ & 1 & 16/3 & $256\times 832$ & $256\times 832$ & concat\_6\_2 \\
			& sum\_6       & -- & --         & (3,3)/3  & $256\times 832$ & $256\times 832$ & mofconv\_6, upsample\_6 \\

			\bottomrule[0.8mm]
		\end{tabular}
	\end{adjustbox}
	\vspace{+2mm}
	\caption{Details of MotionNet.}
	\label{tab_motionnet}
	\vspace{-0mm}
\end{table*}

\section{Details of Architectures}
\label{sec_A}

The detailed architectures of DepthNet, DAM, and MotionNet are specified in \tabref{tab_depthnet}, \tabref{tab_dam}, and \tabref{tab_motionnet}, respectively.
Generic layers, \eg, \emph{conv}, \emph{fc}, \emph{pool}, \etc, are specified starting with a lowercase letter.
Nonlinear activation functions, \eg, \emph{ReLU}, \emph{ELU}, \etc, are abbreviated for visibility.
The encoders of DepthNet and MotionNet are both based on ResNet18~\cite{he2016deep}.
We declare the residual block as \emph{ResBlock} for each encoder.

\noindent\textbf{DepthNet}~
For the decoder of DepthNet, we adopt the structure of MonoDepth2~\cite{godard2019digging}.
DepthNet can predict depth maps with five scales, but we empirically found that single-scale training produces better performance than multi-scale training.
Thus, we predict the depth map from the last layer of the decoder, which is activated with a \emph{sigmoid} function.

\noindent\textbf{MotionNet with DAM}~
For DAM, we design the weighted context and transforming operation for each ego-motion and residual motion feature, respectively.
We attach DAM after the $2^{nd}$, $3^{rd}$, and $4^{th}$ residual layer in the ResNet encoder of MotionNet.
In the case of the ResNet18 encoder, each residual layer is composed of two residual blocks. 
GCNet~\cite{cao2019gcnet} attached the attention module to each residual block, however, we empirically found that deploying DAM only after the residual layer leads to a better trade off between performance and computational time.
%shows the better result considering the computational efficiency.
The details of this ablation is specified in \secref{sec_B} and examples of the spatial attention map are visualized in \secref{sec_C}.
The decoder of MotionNet has multiple refining steps, which has been proposed by Gordon~\etal~\cite{gordon2019depth}.
Every refinement step aggregates the features from the encoding layer and predictions from the lower layer.
Similar to DepthNet, we predict the residual motion field with a single-scale, which is refined from the last layer of the decoder.

\begin{figure}[t] 
	\centering
    \includegraphics[width=0.99\linewidth]{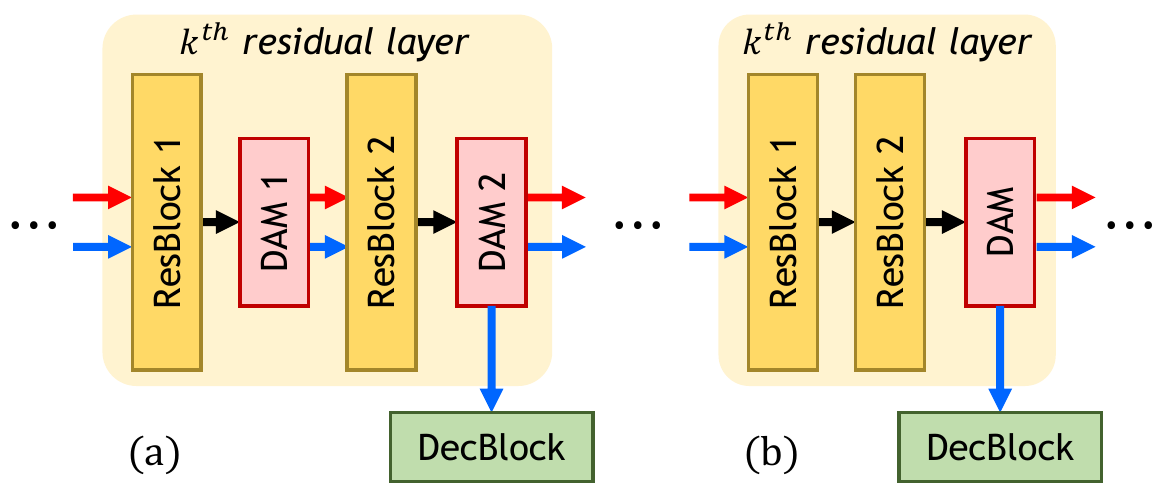}
	\caption{\textbf{Different positional design options for DAM in a residual layer.} We validate two design choices for DAM: (a) after every residual block (GCNet~\cite{cao2019gcnet} style), and (b) after the last residual block in each residual layer.}
	\label{dam_position}
\end{figure}

\begin{table}[t]
\centering
\begin{adjustbox}{width=0.88\linewidth}
\setlength{\tabcolsep}{3mm}
\begin{tabular}{ccccc}
    \Xhline{3\arrayrulewidth}
    \multirow{2}[3]{*}{Models} & \multicolumn{2}{c}{\emph{phase-1}} & \multicolumn{2}{c}{\emph{phase-3}} \\
    \cmidrule(l{2pt}r{2pt}){2-3} \cmidrule(l{2pt}r{2pt}){4-5}
    & \emph{all} & \emph{obj} & \emph{all} & \emph{obj} \\
    \Xhline{2\arrayrulewidth}
    without DAM                 & 0.126 & 0.202 & 0.113 & 0.190 \\
    DAM after every ResBlock    & 0.122 & 0.199 & 0.111 & \textbf{0.181} \\
    DAM after the last ResBlock & \textbf{0.121} & \textbf{0.196} & \textbf{0.109} & 0.182 \\
    \Xhline{3\arrayrulewidth}
\end{tabular}
\end{adjustbox}
\vspace{+2mm}
\caption{\textbf{Ablation study on different positional configurations of DAM.} We follow the same ablation scheme of Sec. 4.2. in our main paper, and measure the AbsRel errors after \emph{phase-1} and \emph{phase-3} on both \emph{all} and \emph{obj} areas. In \emph{phase-3}, we regularize the residual motion field with CSAC.}
\label{tab_dam_position}
\end{table}

\section{Additional Ablation Study}
\label{sec_B}
In this section, we discuss additional ablation studies of different design choices and hyper-parameter settings.
We follow the same configurations of experiments, \eg, dataset and training scheme, which are described in Sec. 4.2. of our main paper.

\begin{figure}[t] 
	\centering
    \includegraphics[width=0.99\linewidth]{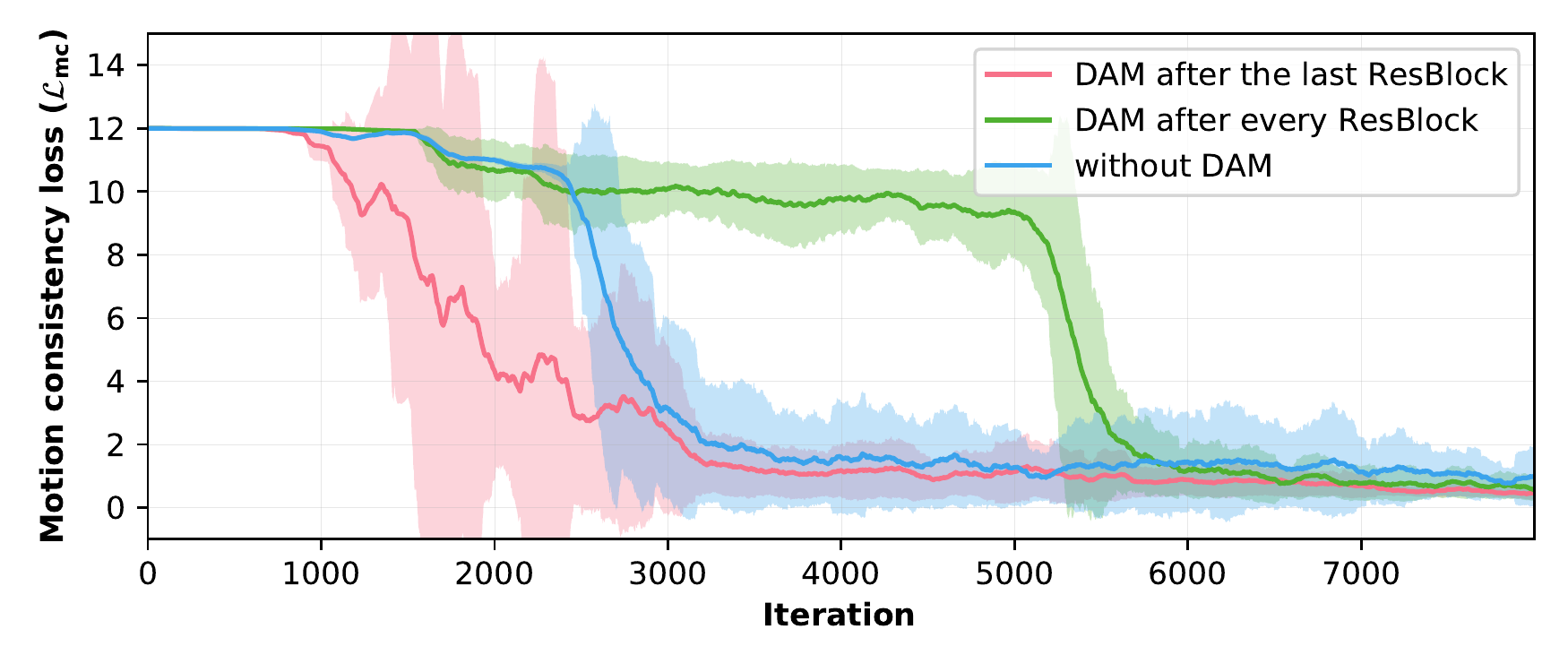}
	\caption{\textbf{Trends of motion field consistency loss in \emph{phase-2} for different DAM configurations.} We train MotionNet on the KITTI dataset from scratch, and plot its training loss.}
	\label{plot_mof_consistency}
\end{figure}

\begin{figure}[t] 
	\centering
    \includegraphics[width=0.99\linewidth]{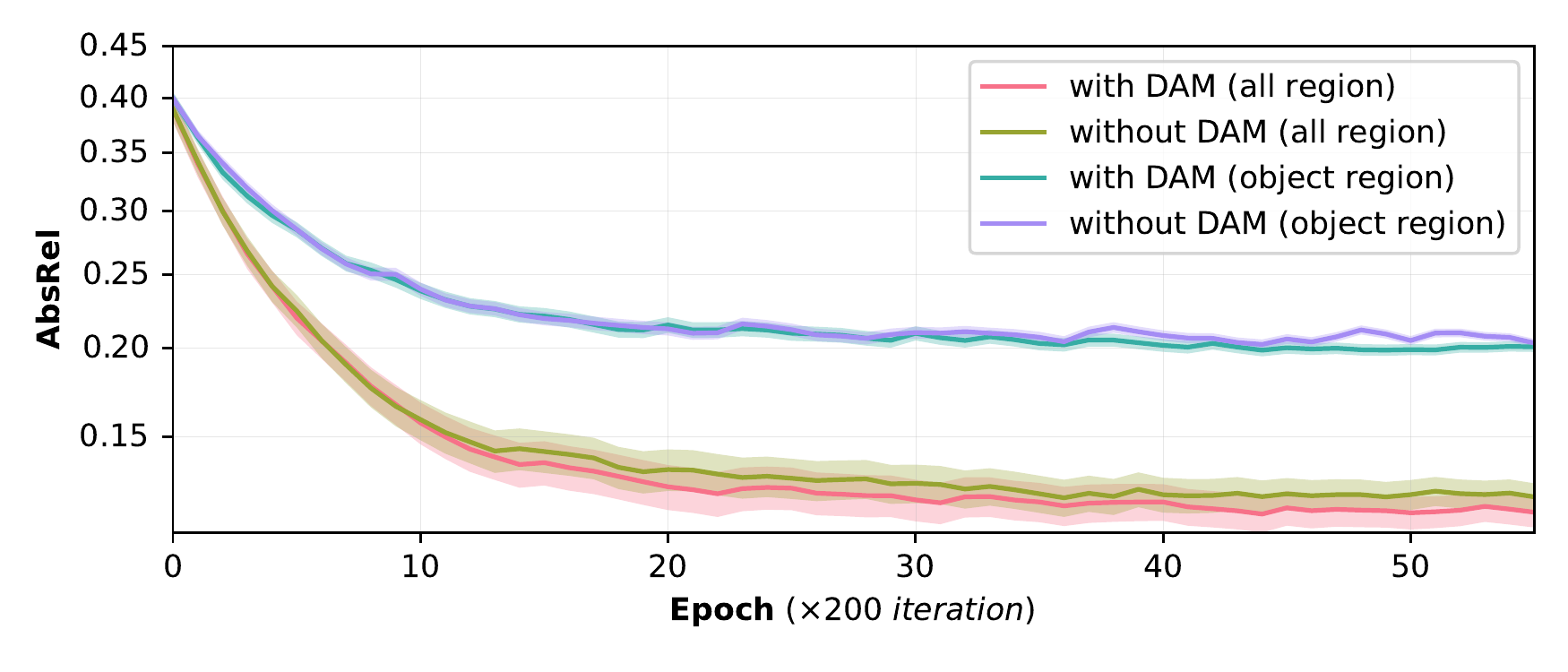}
	\caption{\textbf{Validation error trends in \emph{phase-1} depending on the presence or absence of DAM in the motion encoder.} Models are trained and tested on the KITTI dataset, and we measure the AbsRel errors on both all and object areas.}
	\label{plot_absrel}
\end{figure}

\noindent\textbf{Different positional configurations of DAM}~
As discussed in the previous section, there are two configuration options where to deploy DAM in the encoder of MotionNet: DAM after every residual block, or residual layer.
\figref{dam_position} illustrates each configuration.
In \tabref{tab_dam_position}, we compare them with the performance of monocular depth estimation after \emph{phase-1} and \emph{phase-3}.
The results after \emph{phase-1} show that the configuration of DAM after the last ResBlock produces the lowest error.
However, after \emph{phase-3}, both positional configurations of DAM show similar performance.
Additionally, in \figref{plot_mof_consistency}, we provide the training trends of motion field consistency loss ($\calL_{mc}$) for each DAM configuration: without DAM, DAM after every ResBlock, and DAM after the last ResBlock. 
We follow the training scheme of \emph{phase-2}, however, to see distinctive results, we train MotionNet from random initialization with DepthNet pretrained from \emph{phase-1}.
The results show that the configuration with DAM after the last ResBlock converges fastest.
From these ablation studies, we conjecture that the positional configuration of DAM has a relation with model complexity.
Since the shared motion encoder performs two tasks, there could be a confusion from different self-supervisory signals.
DAM helps to switch the extraction of motion feature in the encoder, however, the convergence will become slow if this switching mechanism is too complicated.
Therefore, considering the training efficiency, we have chosen the configuration of DAM after the last ResBlock.
Finally, we analyse the convergence trends of the networks in \emph{phase-1} with and without DAM in~\figref{plot_absrel} -- applied on the objects only and on the entire image. We can notice a marginal improvement of the depth quality when DAM is employed. It underlines that during this training phase the motion encoder design only has little influence on the depth estimation. However, as underlined by other tests, DAM provides significant improvements on the overall system after the entire training process.

\begin{figure}[t] 
	\centering
    \includegraphics[width=0.77\linewidth]{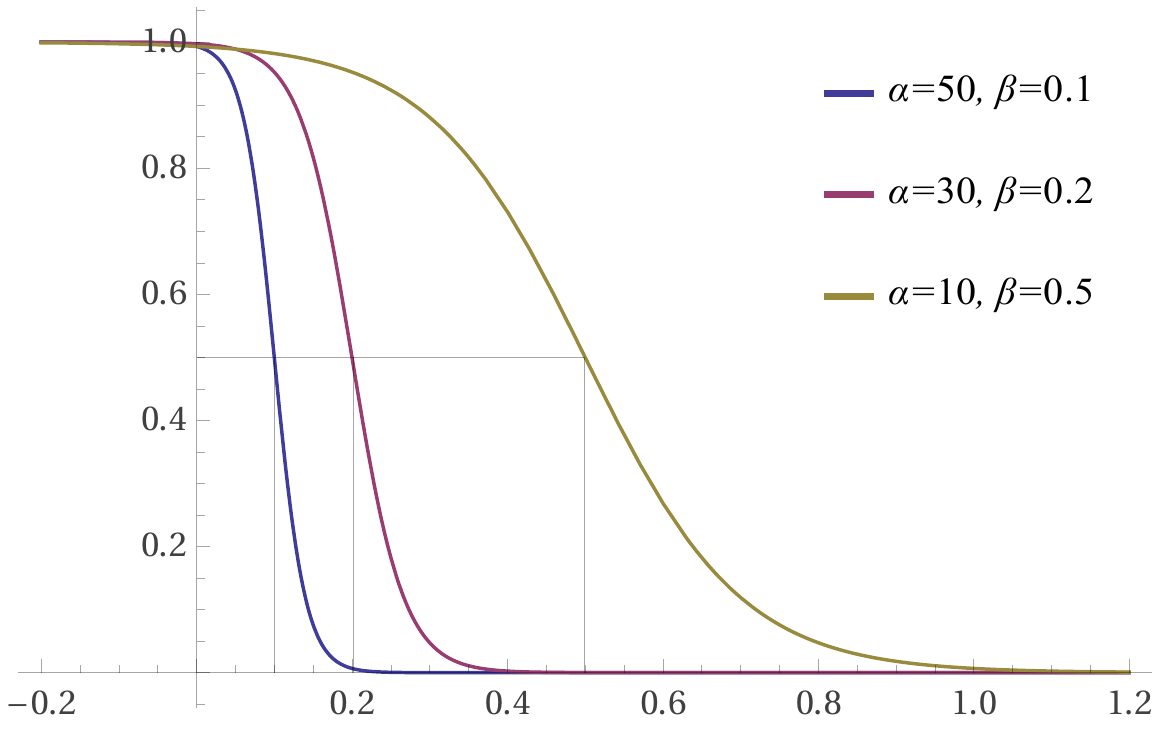}
	\caption{\textbf{Different inlier score mapping with $\alpha$ and $\beta$.} The steeper the curve, the more discretized, so we conduct ablations to find appropriate values.}
	\label{inlier_score}
\end{figure}

\begin{table}[t]
\centering
\begin{adjustbox}{width=0.55\linewidth}
\setlength{\tabcolsep}{5mm}
\begin{tabular}{ccc}
    \Xhline{2\arrayrulewidth}
    $\{\alpha$, $\beta\}$ & D1 (\emph{fg}) & D2 (\emph{fg}) \\
    \Xhline{1\arrayrulewidth}
    \{50, 0.1\} & 34.1 & 41.3 \\
    \{30, 0.2\} & \textbf{32.5} & \textbf{35.7} \\
    \{10, 0.5\} & 33.8 & 37.2 \\
    \Xhline{2\arrayrulewidth}
\end{tabular}
\end{adjustbox}
\vspace{+2mm}
\caption{\textbf{Ablation study of different $\alpha$ and $\beta$ on KITTI Scene Flow 2015 training set.} We set $\alpha = 30$ and $\beta = 0.2$ as our final model of CSAC.}
\label{tab_alpha_beta}
\end{table}

\noindent\textbf{Different $\alpha$ and $\beta$ for inlier score}~
In \figref{inlier_score}, we visualize different mapping of the soft inlier score function $\calF_{inlier}$ according to $\alpha$ and $\beta$ to analyze their impacts. 
The graph shows that a high $\alpha$ value makes the curve sharp, and this leads to increase the discretization of the inlier scores.
In \tabref{tab_alpha_beta}, we conduct an ablation study to find the appropriate values of $\alpha$ and $\beta$.
From the results, we conclude that both high discretization and tolerant mapping have adverse effects on residual motion learning. As a result of this ablation study, we selected $\alpha = 30$ and $\beta = 0.2$ for CSAC.

\begin{figure}[t] 
	\centering
    \includegraphics[width=0.99\linewidth]{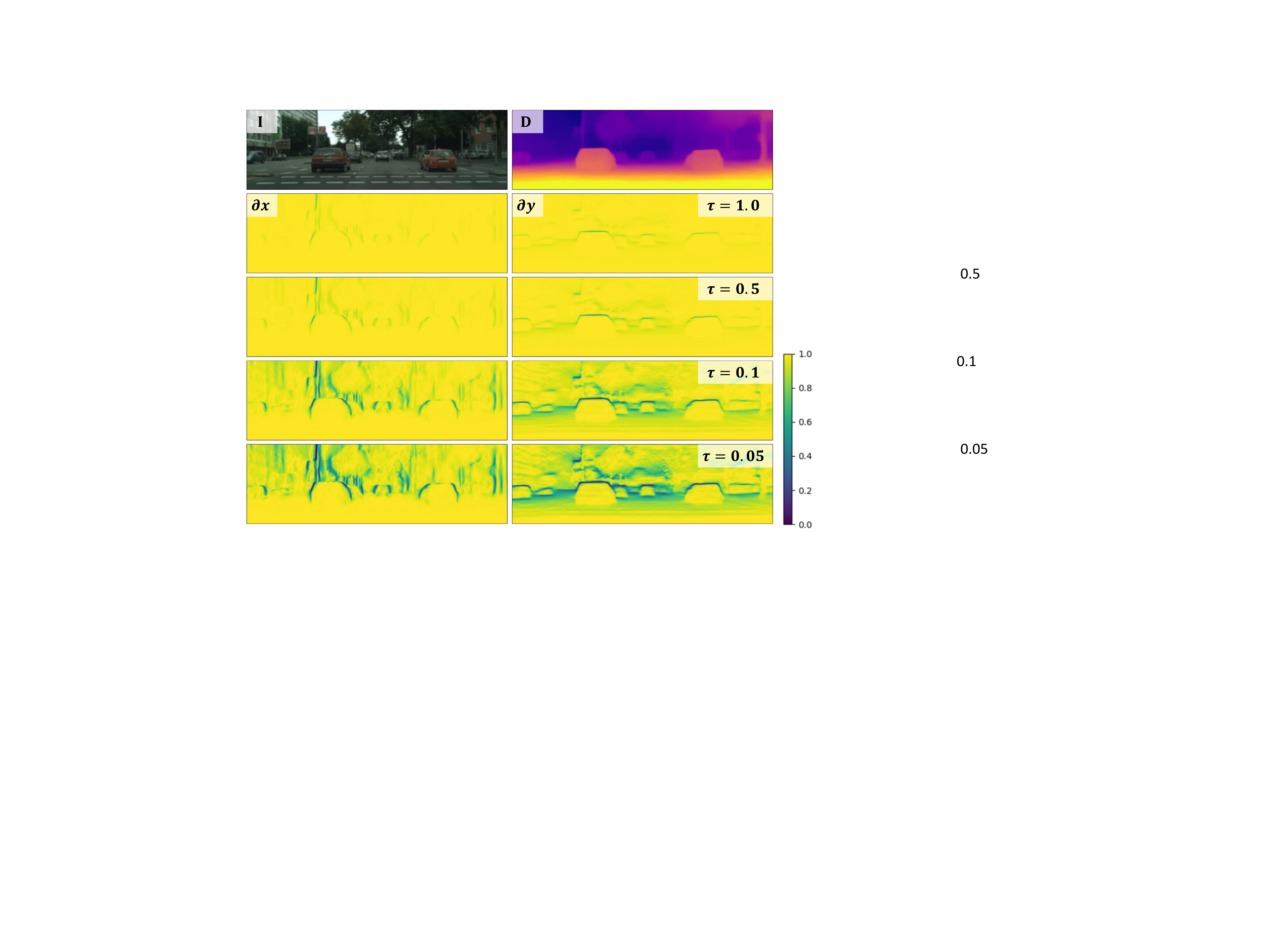}
	\caption{\textbf{Reparameterized gradient maps $e^{-\nabla D / \tau}$ depending on different $\tau$.}  We visualize image, its depth map, and gradients over the $x$ and $y$ axis. Smaller value of $\tau$ %suppress gradients on the edges more strongly.}
	reduces gradients on the edges.}
	\label{smooth_tau}
\end{figure}

\begin{table}[t]
\centering
\begin{adjustbox}{width=0.99\linewidth}
\setlength{\tabcolsep}{4mm}
\begin{tabular}{ccccc}
    \Xhline{2\arrayrulewidth}
     & $\tau = 1.0$ & $\tau = 0.5$ & $\tau = 0.1$ & $\tau = 0.05$ \\
    \Xhline{1\arrayrulewidth}
    Cityscapes-VIS & 0.683 & 0.689 & \textbf{0.712} & 0.704    \\
    \Xhline{2\arrayrulewidth}
\end{tabular}
\end{adjustbox}
\vspace{+2mm}
\caption{\textbf{Motion segmentation results (mean IoU) with different $\tau$.} This table is related (dataset and training scheme) to Table 5 of our main paper.}
\label{tab_smooth_tau}
\end{table}

\noindent\textbf{Reparameterized edge-aware motion smoothness}~
We visualize the reparameterized gradients maps according to different $\tau$ in \figref{smooth_tau}.
The vanilla edge-aware smoothness term ($\tau=1.0$) is not distinctive enough to leverage the boundary prior of objects.
By reparameterizing with a small value of $\tau$, we conjecture that the motion segmentation near the boundary of objects would be improved.
We show motion segmentation results on different $\tau$ in \tabref{tab_smooth_tau}.

\begin{table}[t]
\begin{center}
    \begin{adjustbox}{width=0.99\linewidth}
    \setlength{\tabcolsep}{4mm}
    \begin{tabular}{c|cccc|cccc}
        \Xhline{3\arrayrulewidth}
        Phase & $\lambda_{p}$ & $\lambda_{g}$ & $\lambda_{s}$ & $\lambda_{h}$ & $\lambda_{mr}$ & $\lambda_{ms}$ & $\lambda_{mp}$ & $\lambda_{mc}$ \\
        \Xhline{2\arrayrulewidth}
        \emph{phase-1} & 1.0 & 1.0 & 0.1 & 0.2 & -- & -- & -- & -- \\ 
        \emph{phase-2} & 1.0 & 1.0 & 0.1 & 0.2 & -- & 1.0 & 0.5 & 0.001 \\  
        \emph{phase-3} & 1.0 & 1.0 & 0.1 & 0.2 & 0.2 & 1.0 & 1.0 & 0.001 \\
        \Xhline{3\arrayrulewidth}
    \end{tabular}
    \end{adjustbox}
\vspace{+2mm}
\caption{\textbf{Summarization of loss weights for each phase.} }
\label{tab_loss_weight}
\end{center}
\end{table}

\begin{table}[t]
\centering
\begin{adjustbox}{width=0.99\linewidth}
\setlength{\tabcolsep}{3mm}
\begin{tabular}{cccccccccc}
    \Xhline{3\arrayrulewidth}
    \multirow{2}[3]{*}{\shortstack{$\lambda_{mr}$ \\ $\lambda_{ms}$}} & \multicolumn{3}{c}{0.1} & \multicolumn{3}{c}{0.2} & \multicolumn{3}{c}{0.3} \\
    \cmidrule(l{2pt}r{2pt}){2-4} \cmidrule(l{2pt}r{2pt}){5-7} \cmidrule(l{2pt}r{2pt}){8-10}
    & 0.1 & 0.5 & 1.0 & 0.1 & 0.5 & 1.0 & 0.1 & 0.5 & 1.0 \\
    \Xhline{2\arrayrulewidth}
    \emph{all} & 0.123 & 0.119 & 0.116   & 0.116 & \textbf{0.113} & \textbf{0.113}   & 0.119 & 0.123 & 0.128 \\
    \emph{obj} & 0.212 & 0.205 & 0.194   & 0.196 & 0.190          & \textbf{0.188}   & 0.206 & 0.217 & 0.224 \\
    \Xhline{3\arrayrulewidth}
\end{tabular}
\end{adjustbox}
\vspace{+2mm}
\caption{\textbf{Ablation study of different loss weights, $\lambda_{mr}$ and $\lambda_{ms}$, on the KITTI dataset.} We measure the AbsRel errors on both \emph{all} and \emph{obj} areas.}
\label{tab_ablation_weight}
\end{table}

\noindent\textbf{Configuration of loss weights}~
We summarize the learning parameters for each dataset in \tabref{tab_loss_weight}.
The photometric loss, $\calL_p$, is defined as follows:
\begin{equation}
\Scale[0.99]
{
\begin{aligned}
\calL_p =
\sum \left ( \gamma_1 \left | {\mathbf{I} - \hat{\mathbf{I}} } \right |_1 + \gamma_2 ( 1 - SSIM(\mathbf{I}, \hat{\mathbf{I}}) )  \right ) ,
\end{aligned}
}
\label{eq_loss_p}
\end{equation}
where $SSIM$ is the structural similarity loss~\cite{wang2004image}, and \{$\gamma_1$, $\gamma_2$\} is set to \{0.3, 1.5\} based on cross-validation. 
Other parameters of our loss functions are the same as previous works as described in Sec. 3.4. of our main paper.

\noindent\textbf{Different loss weights of $\lambda_{mr}$ and $\lambda_{ms}$}~
%We conduct ablation studies to verify $\lambda_{mr}$ and $\lambda_{ms}$ on the KITTI ablation.
To justify the parametrization of the hyper-parameters $\lambda_{mr}$ and $\lambda_{ms}$, we propose another ablation study conducted on the KITTI dataset.
For this experiment, we follow the training scheme and dataset described in Sec. 4.2. of our main paper.
As provided in \tabref{tab_ablation_weight}, our motion regularization with CSAC shows stable training with $\lambda_{mr} = 0.2$ and $\lambda_{ms} = 1.0$.

\begin{table*}[t]
\centering
\vspace{-0mm}
\begin{adjustbox}{width=0.94\textwidth}
\setlength{\tabcolsep}{2mm}
\begin{tabular}{lccccccccccc}

\Xhline{4\arrayrulewidth}
\multirow{2}{*}{Method} & \multirow{2}{*}{Phase} & \multirow{2}{*}{Training} & \multirow{2}{*}{Test} & \multicolumn{4}{c}{Error metric $\downarrow$} & & \multicolumn{3}{c}{Accuracy metric $\uparrow$}  \\
\cline{5-8} \cline{10-12}
 &  &  &  & AbsRel  & SqRel & RMSE & RMSE log & & $\delta < 1.25$ & $\delta < 1.25^2$ & $\delta < 1.25^3$ \\ \Xhline{2\arrayrulewidth}

Baseline        & \emph{phase-1} & K & K & 0.126 & 0.975 & 5.244 & 0.211 &  & 0.849 & 0.946 & 0.979 \\
Ours (DAM)      & \emph{phase-1} & K & K & 0.123 & 0.920 & 5.212 & 0.205 &  & 0.854 & 0.945 & 0.978 \\
Ours (DAM)      & \emph{phase-3} & K & K & 0.120 & 0.895 & 4.972 & 0.196 &  & 0.859 & 0.950 & 0.980 \\
Ours (DAM+CSAC) & \emph{phase-3} & K & K & \textbf{0.114} & \textbf{0.876} & \textbf{4.715} & \textbf{0.191} &  & \textbf{0.872} & \textbf{0.955} & \textbf{0.981} \\
\Xhline{1\arrayrulewidth}

Baseline        & \emph{phase-1} & C+K & K & 0.122 & 0.907 & 4.985 & 0.207 &  & 0.862 & 0.953 & 0.980 \\
Ours (DAM)      & \emph{phase-1} & C+K & K & 0.119 & 0.883 & 5.021 & 0.206 &  & 0.861 & 0.954 & 0.979 \\
Ours (DAM)      & \emph{phase-3} & C+K & K & 0.116 & 0.845 & 4.790 & 0.194 &  & 0.868 & 0.957 & 0.979 \\
Ours (DAM+CSAC) & \emph{phase-3} & C+K & K & \textbf{0.111} & \textbf{0.805} & \textbf{4.708} & \textbf{0.187} &  & \textbf{0.875} & \textbf{0.962} & \textbf{0.981} \\
\Xhline{1\arrayrulewidth}

Baseline        & \emph{phase-1} & C & C & 0.128 & 1.322 & 6.942 & 0.198 &  & 0.833 & 0.949 & 0.978 \\
Ours (DAM)      & \emph{phase-1} & C & C & 0.127 & 1.330 & 6.903 & 0.196 &  & 0.838 & 0.950 & 0.979 \\
Ours (DAM)      & \emph{phase-3} & C & C & 0.124 & 1.281 & 6.818 & 0.189 &  & 0.849 & \textbf{0.951} & 0.981 \\
Ours (DAM+CSAC) & \emph{phase-3} & C & C & \textbf{0.116} & \textbf{1.213} & \textbf{6.695} & \textbf{0.186} &  & \textbf{0.852} & \textbf{0.951} & \textbf{0.982} \\
\Xhline{1\arrayrulewidth}

Baseline        & \emph{phase-1} & W & W & 0.161 & 1.724 & 7.825 & 0.217 &  & -- & -- & -- \\
Ours (DAM)      & \emph{phase-1} & W & W & 0.159 & 1.707 & 7.816 & 0.215 &  & -- & -- & -- \\
Ours (DAM)      & \emph{phase-3} & W & W & 0.155 & 1.692 & 7.448 & 0.212 &  & -- & -- & -- \\
Ours (DAM+CSAC) & \emph{phase-3} & W & W & \textbf{0.148} & \textbf{1.686} & \textbf{7.420} & \textbf{0.210} &  & -- & -- & -- \\
\Xhline{4\arrayrulewidth}

\end{tabular}
\end{adjustbox}
\vspace{+2mm}
\caption{\textbf{Full results of our models for monocular depth estimation on KITTI (K) Eigen test set, Cityscapes (C) test set, and Waymo Open Dataset (W).} The models pretrained on Cityscapes and fine-tuned on KITTI are denoted by `C+K'. For each partition, \textbf{Bold}: Best.}
\label{tab_kitti_supp}
\end{table*}

\section{Additional Experimental Results}
\label{sec_C}
In this section, we present additional experimental results as an extension of Sec. 4. of our main paper.

\noindent\textbf{Full table of monocular depth estimation}~
We provide the full results of our models for the task of monocular depth estimation in \tabref{tab_kitti_supp}. 
The results consistently show that our proposed modules, DAM and CSAC, favorably work on three different dataset: KITTI, Cityscapes, and Waymo Open Dataset.

% \begin{figure}[t] 
% 	\centering
%     \includegraphics[width=0.99\linewidth]{figure/view_synthesis}
% 	\caption{\textbf{Examples of 3D view synthesis.} We generate novel target view synthesis using predicted depth and pixel-wise motion field.
% 	%We demonstrate target view synthesis using predicted depth and pixel-wise motion field.
% 	}
% 	\label{view_synthesis}
% \end{figure}

\begin{figure*}[t] 
	\centering
    \includegraphics[width=0.94\linewidth]{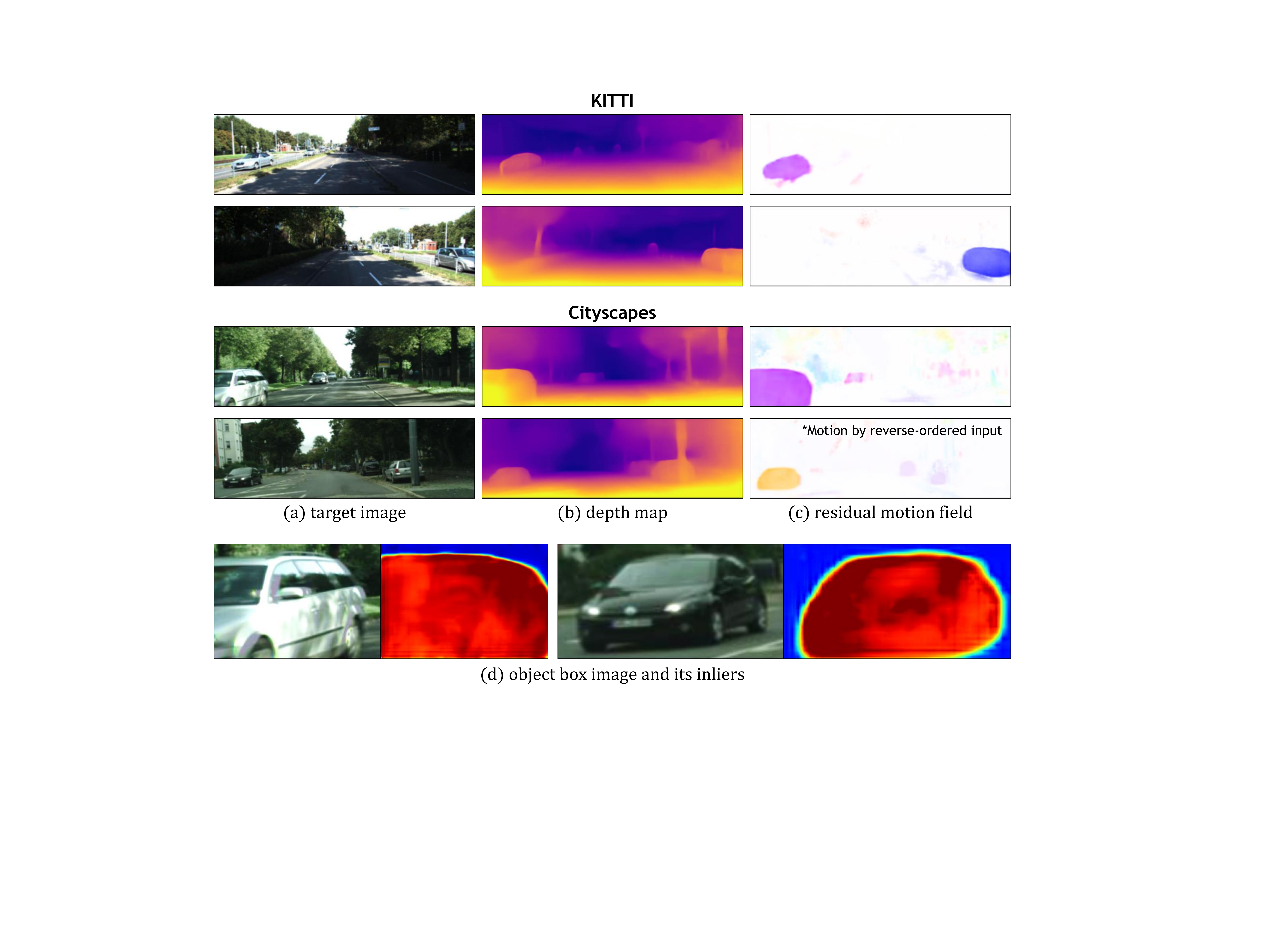}
	\caption{\textbf{Qualitative results of depth and residual motion estimation in KITTI and Cityscapes.} The residual motion field is mapped into the HSV color space.
	}
	\label{qualitative_supp}
\end{figure*}

\begin{figure*}[t] 
	\centering
    \includegraphics[width=0.99\linewidth]{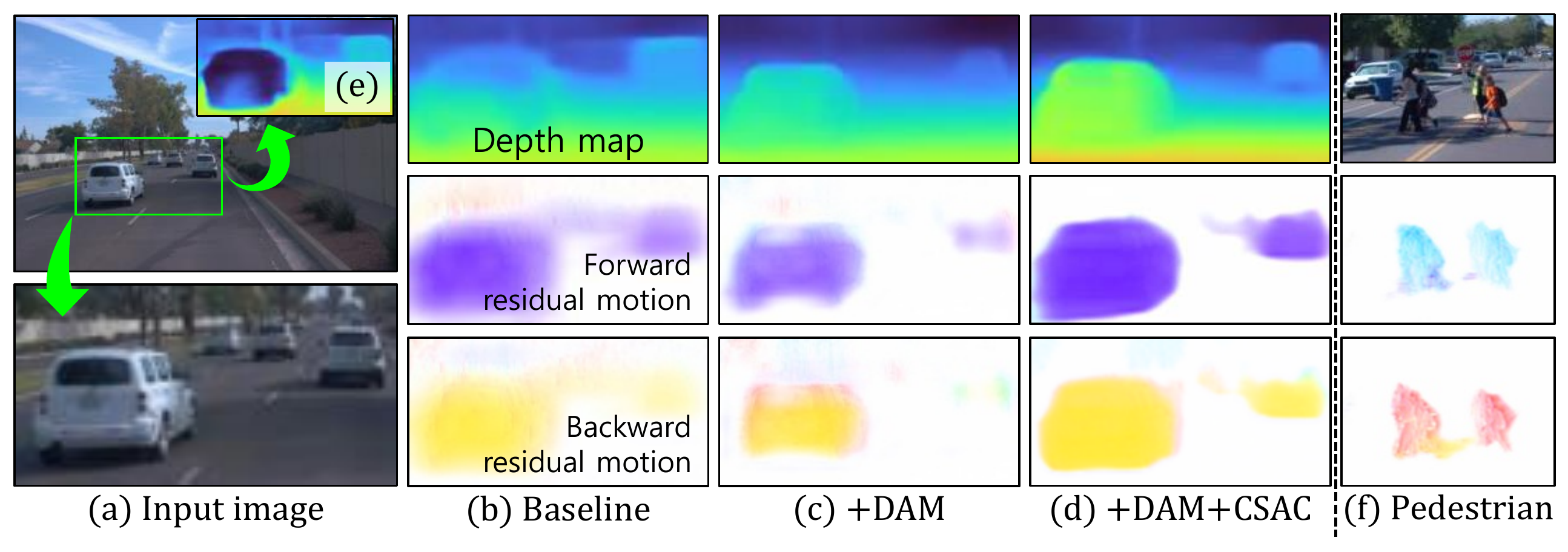}
	\caption{\textbf{Qualitative results of depth map and bidirectional residual motion field in Waymo Open Dataset.} AbsRel errors on \emph{all} / \emph{obj}: (b) 0.154 / 0.328, (c) 0.149 / 0.250 (improved distinction between \emph{obj} and background), (d) 0.135 / 0.164 (sharpen object boundary). (e) Result of a diverged depth map, if auto-masking proposed by MonoDepth2~\cite{godard2019digging} fails. (f) Results of residual motion field for pedestrians.
	}
	\label{qualitative_waymo}
\end{figure*}

\begin{figure*}[t] 
	\centering
    \includegraphics[width=0.99\linewidth]{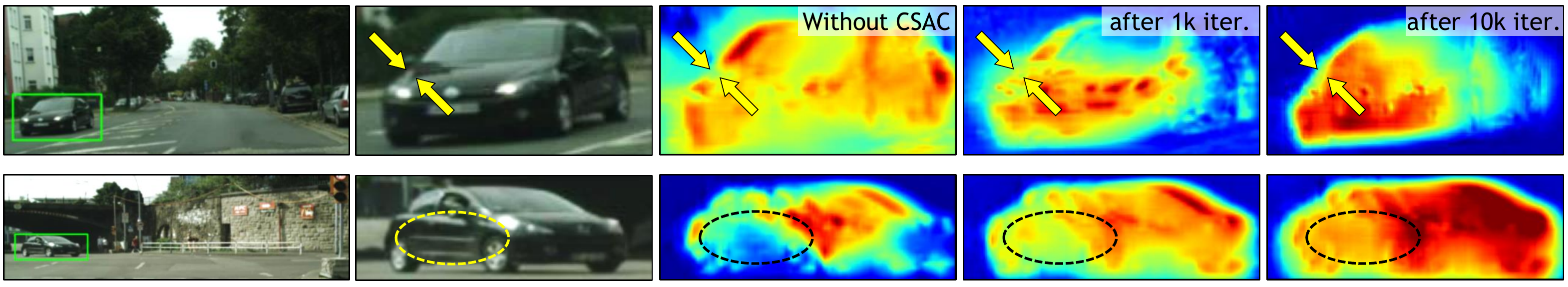}
	\caption{\textbf{Qualitative results of motion inliers in Cityscapes.} CSAC makes the motion boundaries more clear and sharper as shown in the first row, and the motion holes in homogeneous regions more consistent on the rigid objects as demonstrated in the second row.
	}
	\label{qualitative_csac_supp}
\end{figure*}

\noindent\textbf{Qualitative results}~
% In order to show the visual effect of the regularization by CSAC, we demonstrate a novel view synthesis with predicted depth map and motion field.
% In \figref{view_synthesis}, three scenes from the Cityscapes test set are synthesized with 3D point cloud.
%
%Our module consistently transforms the whole points on moving objects,
In addition to Fig. 7 in our main paper, we visualize the depth map and residual motion field in \figref{qualitative_supp} and \figref{qualitative_waymo}.
\figref{qualitative_csac_supp} shows two representative effects of the regularization through CSAC: sharpen boundaries of object's motion, and motion hole filling in the homogeneous areas.
Our module consistently preserves the shape of the moving objects, while the baseline model distorts the appearance of objects.

% Interestingly, DAM tries to focus on static regions such as driving roads to estimate camera ego-motion. 

{\small
\bibliographystyle{ieee_fullname}
\bibliography{__egbib}
}

% \input{__doc_supp}
% \end{appendices}

\end{document}